\documentclass{article}

\usepackage[preprint]{neurips_2025}

\usepackage[utf8]{inputenc} 
\usepackage[T1]{fontenc}    
\usepackage{hyperref}       
\usepackage{url}            
\usepackage{booktabs}       
\usepackage{amsfonts}       
\usepackage{nicefrac}       
\usepackage{microtype}      
\usepackage{xcolor}         
\usepackage{amsmath, amssymb}
\usepackage{graphicx}
\usepackage{afterpage}
\usepackage{wrapfig}
\usepackage{subfig}
\usepackage{caption}
\usepackage{multirow}
\usepackage{colortbl}
\newtheorem{definition}{Definition}

\newcommand{\tablestyle}[2]{\setlength{\tabcolsep}{#1}
\renewcommand{\arraystretch}{#2}\centering\scriptsize}
\renewcommand{\paragraph}[1]{\vspace{1.25mm}\noindent\textbf{#1}}

\newcolumntype{x}[1]{>{\centering\arraybackslash}p{#1pt}}
\newcolumntype{y}[1]{>{\raggedright\arraybackslash}p{#1pt}}
\newcolumntype{z}[1]{>{\raggedleft\arraybackslash}p{#1pt}}

\newcommand{\app}{\raise.17ex\hbox{$\scriptstyle\sim$}}

\definecolor{deemph}{gray}{0.6}

\definecolor{baselinecolor}{gray}{.9}

\definecolor{dt}{HTML}{ADCAD8}
\definecolor{dt2}{HTML}{cddfe7}

\definecolor{defaultcolor}{HTML}{E8E2F7}

\let\cite\citep

\renewcommand{\paragraph}[1]{\vspace{1.25mm}\noindent\textbf{#1}}
\newlength\savewidth

\newcolumntype{x}[1]{>{\centering\arraybackslash}p{#1pt}}
\newcolumntype{y}[1]{>{\raggedright\arraybackslash}p{#1pt}}
\newcolumntype{z}[1]{>{\raggedleft\arraybackslash}p{#1pt}}
\definecolor{degray}{gray}{.6}

\title{HyperNAS: Enhancing Architecture Representation for NAS Predictor via Hypernetwork}

\author{
Jindi Lv$^{1}$\quad 
Yuhao Zhou$^{1}$\quad
Yuxin Tian$^{1}$\quad 
\textbf{Qing Ye}$^{1}$\quad
\textbf{Wentao Feng}$^{1}$\quad
\textbf{Jiancheng Lv}$^{1}$\quad
\\
$^{1}$School of Computer Science, Sichuan University\;
}

\begin{document}

\maketitle

\begin{abstract}
Time-intensive performance evaluations significantly impede progress in Neural Architecture Search (NAS). 
To address this, neural predictors leverage surrogate models trained on proxy datasets,  allowing for direct performance predictions for new architectures.
However, these predictors often exhibit poor generalization due to their limited ability to capture intricate relationships among various architectures. 
In this paper, we propose HyperNAS, a novel neural predictor paradigm for enhancing architecture representation learning. 
HyperNAS consists of two primary components: a global encoding scheme and a shared hypernetwork. The global encoding scheme is devised to capture the comprehensive macro-structure information, while the shared hypernetwork serves as an auxiliary task to enhance the investigation of inter-architecture patterns. 
To ensure training stability, we further develop a dynamic adaptive multi-task loss to facilitate personalized exploration on the Pareto front. 
Extensive experiments across five representative search spaces, including ViTs, demonstrate the advantages of HyperNAS, particularly in few-shot scenarios. For instance, HyperNAS strikes new \textit{state-of-the-art} results, with 97.60\% top-1 accuracy on CIFAR-10 and 82.4\% top-1 accuracy on ImageNet, using at least 5.0$\times$ fewer samples.
\end{abstract}

\section{Introduction}
Neural architecture search (NAS) automates the design of neural networks and surpasses human-designed architectures in various tasks~\cite{howard2019searching,wang2020fcos,sarah2024llama}, particularly on ImageNet~\cite{chen2021neural}, diverse and less-studied datasets~\cite{shen2022efficient}, and in hardware-constrained settings~\cite{cai2018proxylessnas}. 

In practice, NAS is a bi-level optimization problem that demands substantial resources~\cite{zoph2016neural,real2019regularized}. 
The whole process involves optimizing search strategy and performance evaluation, with the latter often requiring thousands of full network training~\cite{white2023neural}.
As a remedy, one-shot NAS~\cite{liu2018darts,cai2018proxylessnas,chu2020fair} and predictor-based NAS~\cite{luo2018neural,wen2020neural,chen2021contrastive,lu2023pinat,yi2023nar,ji2024cap} are proposed.
Specifically, one-shot NAS methods involve training a single supernet that encompasses all potential architectures and evaluating their performance by directly inheriting weights from the supernet. 
However, they suffer from optimization gap issues due to the weight-sharing mechanism~\cite{xie2021weight, chu2020fair, song2024efficient}. 
In contrast, predictor-based NAS methods offer a promising alternative. 
They use a proxy dataset to learn an accurate mapping between architectures and their associated accuracies, enabling rapid performance predictions for unseen architectures. 
Despite notable advancements, we argue that predictor-based NAS still suffers from the following limitations: \textit{isolated cell encoding}~\cite{wen2020neural,lu2023pinat, akhauri2024encodings} and \textit{poor generalization}~\cite{chen2021contrastive,yan2021cate, yan2020does}.

Firstly, in cell-based search spaces, prior works often rely on an isolated cell to estimate architecture performance, assuming the architecture is composed of identical cells arranged in a handcrafted hierarchical structure~\cite{pham2018efficient,liu2018darts,ying2019bench,dong2020bench}.
While this simplification effectively trims computational budgets, it neglects essential macro-structural insights like reduction
operations between cells and inter-cell dependencies~\cite{white2023neural}. 
Secondly, neural predictors are trained on small proxy datasets to excavate intricate relationships among architectures, enabling robust generalization for accurately assessing unseen architectures~\cite{wen2020neural,jing2022graph,lu2023pinat}.  
However, the architecture relationship within the search space is often highly complex and non-linear, making it difficult to capture its underlying patterns with simple predictors alone.

In this work, we propose HyperNAS, a novel neural predictor paradigm for enhancing architecture representation learning.
To address the first limitation, we introduce a global architecture encoding scheme that effectively captures macro-structural information by seamlessly integrating features from each cell.
To tackle the second challenge, we incorporate a hypernetwork as an auxiliary task to generate weights for diverse architectures,  leveraging its adaptive nature to handle architectural variability~\cite{chauhan2023brief}.
By jointly optimizing the neural predictor and hypernetwork, this approach enables the shared architecture encoder to gain deeper insights into underlying architectural patterns, mitigating the risk of overfitting to architecture-accuracy pairs.
Furthermore, we propose a dynamic adaptive multi-objective loss function with a preference coefficient, enabling personalized exploration on the Pareto front while enhancing training stability and performance. Unlike prior methods~\cite{liu2021homogeneous,xie2023architecture}, HyperNAS eliminates the need to create augmentation data pairs for training.


Our contributions are summarized as follows:
\begin{itemize}
    \item We propose HyperNAS, a neural predictor paradigm designed to enhance architecture representation learning with limited training samples. To the best of our knowledge, this is the first work focusing on improving generalization by incorporating an auxiliary task.
    \item HyperNAS boosts architecture representations by integrating a global encoding scheme and a shared hypernetwork acting as an auxiliary task. The global encoding scheme captures the macro-structural context, while the shared hypernetwork uncovers underlying inter-architecture relationships.
    \item We design a dynamic adaptive multi-task loss function that enables personalized exploration on the Pareto front, ensuring training stability and balanced optimization.
    \item Extensive experiments across five representative NAS search spaces demonstrate the superior generalization capability of HyperNAS in few-shot scenarios. 
\end{itemize}

\section{Related Work}
\label{headings}
\subsection{Performance Evaluation of NAS}
NAS automates the design of neural networks, aiming to discover architectures that outperform manually crafted ones. 
Prior NAS methods often relied on reinforcement learning~\cite{zoph2016neural} and evolutionary algorithms~\cite{real2019regularized, lyu2021multiobjective}. 
Despite being competitive, these methods required substantial computational resources.
To tackle this challenge, one-shot NAS methods are developed~\cite{cai2018proxylessnas,movahedi2022lambda,zhang2023shiftnas}.
These methods involve training a single supernet to share the parameters across sub-networks, drastically eliminating computation intensity. 
However, one-shot NAS methods often face an optimization gap issue~\cite{xie2021weight}, potentially impacting their effectiveness. To overcome this limitation, predictor-based NAS methods have been promising solutions.

Predictor-based NAS methods leverage surrogate models to estimate the performance of candidate architectures, enabling efficient and effective evaluations. 
Given limited training samples, prior works have focused on improving the generalization of neural predictors through two principal aspects: \textit{data augmentation} and \textit{representation enhancement}.
Data augmentation methods commonly employ techniques such as synthetic data generation~\cite{liu2021homogeneous}, transfer learning~\cite{liu2022bridge}, and semi-supervised learning~\cite{tang2020semi}. 
For instance, HAAP~\cite{liu2021homogeneous} employs the permutation invariance of graph-format architecture to synthetic homogeneous representations for training data. 
Moreover, Semi-NAS~\cite{tang2020semi} employs semi-supervised learning to combine a small amount of labeled data with a larger pool of unlabeled data, enhancing the predictor generalization abilities. 
In contrast, representation enhancement methods focus on designing architecture encoders, such as MLPs~\cite{liu2018progressive,white2021bananas,xu2021renas}, GCNs~\cite{li2020neural,wen2020neural,shi2020bridging,chen2021contrastive}, GINs~\cite{yan2020does}, and Transformers~\cite{lu2021tnasp,lu2023pinat}, aiming to provide more informative architecture representations to improve prediction accuracy.
Despite significant progress, neural predictors still face two limitations.
Firstly, in the cell-based search space, they often use an individual cell to evaluate architecture performance, missing valuable information from the marco structure. 
Secondly, they often generalize poorly due to their inability to capture complex architecture relationships.

To tackle the aforementioned limitations, we propose a novel predictor paradigm named HyperNAS.
It addresses the first limitation by devising a global architecture encoding scheme, ensuring the integrity of architectural representations. 
Additionally, HyperNAS incorporates a shared hypernetwork as an auxiliary task to generate weights for various architectures accordingly. Dynamic inputs help this shared hypernetwork to introduce a new level of adaptability~\cite{chauhan2023brief}, allowing for a more comprehensive exploration of architectural relationships.

\subsection{Hypernetwork}
The hypernetwork is a specialized neural network designed to generate weights for a target network~\cite{ha2017hypernetworks}. 
This approach allows for efficient parameter sharing and adaptation, making hypernetwork a powerful tool in various domains, including meta-learning~\cite{zhao2020meta,beck2023hypernetworks,cho2024hypernetwork}, continual learning~\cite{von2019continual,chandra2023continual,hemati2023partial}, and generative modeling~\cite{ratzlaff2019hypergan,schurholt2022hyper,do2020structural}.

Hypernetworks have also been applied to the NAS domain, with two methods currently available: SMASH~\cite{brock2017smash} and GHN~\cite{zhang2018graph}. 
Both methods utilize hypernetworks to dynamically generate weights for various candidate architectures, enabling efficient exploration of the architecture space. 
By employing well-trained hypernetworks as architecture performance evaluators, they significantly accelerate the search process. 
However, their ranking ability is constrained by the hypernetwork's optimization objective, potentially leading to suboptimal rankings and reduced search quality.

In this paper, we shift away from using the hypernetwork directly as an architecture assessor. Instead, we use it as an auxiliary task to complement neural predictors excelling in ranking performance. 
This dual-task approach combines the strengths of hypernetworks and neural predictors to better model architectural relationships and improve generalization.

\begin{figure*}[t!]
    \centering
    \includegraphics[width=0.95\textwidth]{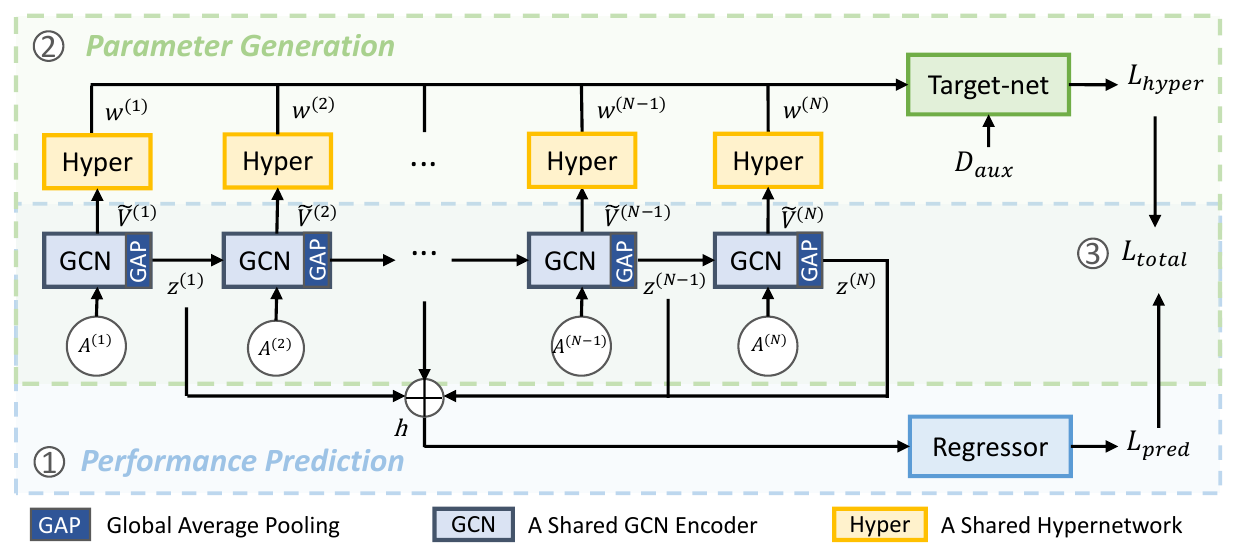}
    \caption{Overview of HyperNAS. HyperNAS is a multi-task paradigm designed to enhance architecture representation learning. It consists of two tasks: the architecture performance prediction task (blue dashed box) and the architecture parameter generation task (green dashed box). The GCN encoder serves as a shared backbone across both tasks.}
    \label{fig:HyperNAS_overview}
    \vspace{-15pt}
\end{figure*}

\section{Methodology}
\label{sec:methodology}
\subsection{Architecture Space}
In the context of a cell-based NAS search space, we define each architecture as a collection of cell architectures $\mathcal{A}=\{\mathcal{A}^{(i)}\}_{i=1}^N$, where $\mathcal{A}^{(i)}$ represents the architecture of the $i$-th cell. 
Each cell is modeled as a directed acyclic graph (DAG), denoted as $\mathcal{A}^{(i)}=\{\mathcal{E}^{(i)}, \mathcal{V}^{(i)}\}$. Here, $\mathcal{E}^{(i)}\in R^{F\times F}$ is an upper triangular adjacency matrix describing the connections between nodes in the $i$-th cell, while $\mathcal{V}^{(i)}\in R^{F\times d}$ represents the node features in the $i$-th cell, with $F$ being the number of nodes and $d$ the dimension of each node feature vector.
\subsection{Overall Framework}
Figure \ref{fig:HyperNAS_overview} illustrates the overall framework of our proposed HyperNAS, which integrates a neural predictor and a shared hypernetwork to jointly facilitate a comprehensive understanding of architecture relationships. 
Initially, we sequentially feed each cell architecture $\mathcal{A}^{(i)}$ into a shared GCN encoder to produce the associated node features $\widetilde{\mathcal{V}}^{(i)}$ and cell features $z^{(i)}$.

For the neural predictor, we propose a global architecture encoding scheme that aggregates the cell features $z^{(i)}$ to derive a global feature $h$. 
By feeding this global feature $h$ into an accuracy regressor, the predictor loss $\mathcal{L}_{pred}$ is computed with the yielded predicted accuracy $\hat{y}$ and the ground-truth label $y$.
In parallel, the shared hypernetwork takes node features $\widetilde{\mathcal{V}}^{(i)}$ to progressively generate weights $w^{(i)}$ for each cell. 
These weights $\{w^{(i)}\}_{i=1}^N$ are then applied to the given architecture, enabling the calculation of the target network loss $\mathcal{L}_{hyper}$ using the auxiliary data $\mathcal{D}_{aux}$. 
To ensure training stability, we propose a dynamic adaptive multi-task loss function with a preference coefficient, ensuring convergence towards a personalized solution on the Pareto front. 

After training with $M$ different architecture-accuracy pairs, HyperNAS can robustly predict accuracy and directly generate promising initial parameters for an unseen architecture through only a single forward inference. Notably, HyperNAS can selectively close task branches during evaluation based on practical scenarios. 
In this paper, we primarily focus on neural predictors, deactivating the hypernetwork branch during inference when no specific requirements are present.

\subsection{Global Architecture Encoding}
In NAS, using individual cell representations to evaluate architectures overlooks critical macro-structural contexts, such as specific reduction operations and intricate cell interactions, which are essential for determining architecture performance. 
To address this issue, we propose a global architecture encoding scheme for neural predictors, as illustrated in Figure \ref{fig:HyperNAS_overview} with the blue dashed box, which offers a holistic view of the architecture.

In this study, we use a GCN tailored for DAGs, as implemented in NP~\cite{wen2020neural}, to serve as a shared architecture encoder, denoted as $G$. 
The GCN iteratively updates the node features through the following formulation:
\begin{equation}
    \mathcal{V}_{l+1}=\frac{1}{2}\mathrm{ReLU}(\mathcal{E}\mathcal{V}_lW_l^+)+\frac{1}{2}\mathrm{ReLU}(\mathcal{E}^T\mathcal{V}_lW_l^-)
\end{equation}
Here, $\mathcal{V}_{l}$ represents the node features at layer $l$, $\mathcal{E}$ is the adjacency matrix, and $W_l^+\in R^{d\times d}$ and $W_l^-\in R^{d\times d}$ are the trainable weight matrices for forward and backward information flows at laye $l$. 

To achieve global encoding, we employ this shared encoder to sequentially integrate each cell into the encoding process. 
Importantly, before being fed into the GCN encoder, the node features $\mathcal{V}^{(i)}$ for each cell should be combined with the preceding cell features $z^{(i-1)}$, as formulated below:
\begin{equation}
    \mathcal{V}^{(i)}=\{v+z^{(i-1)}|v\in \mathcal{V}^{(i)}\}_{i=1}^{N}
\end{equation}
Each cell features $z^{(i)}\in \mathrm{R}^{1\times d}$ encapsulate the overall context from preceding cells and are derived by pooling corresponding GCN output $\widetilde{\mathcal{V}}^{(i)}\in \mathrm{R}^{F\times d}$:
\begin{equation}
    z^{(0)}=0
\end{equation}
\begin{equation}
    z^{(i)}=\mathrm{pool}(\widetilde{\mathcal{V}}^{(i)}),\forall i= 1,\cdots,N
\end{equation}
By incorporating these cell features, the shared GCN encoder can easily distinguish between cells with the same architecture. We also have tried to feed cell position embeddings instead of preceding cell features $z^{(i-1)}$ into the GCN encoder,  which showcases unsatisfying performance. 

After encoding all cells, we derive a global architecture embeddings $h$: 
\begin{equation}
    h=\frac{1}{N}\sum_{i=1}^N z^{(i)}
\end{equation}
The global embeddings incorporate comprehensive macro-structure information of the architecture, which is subsequently utilized by a regressor $f_\theta$ to predict the performance 
$\hat{y}$ of the architectures:
\begin{equation}
    \hat{y}=f_\theta(h)
\end{equation}
The loss for neural predictor $\mathcal{L}_{pred}$ is then calculated using the mean squared error (MSE). 

\subsection{Hypernetwork Auxiliary Task}
Neural predictors often struggle with poor generalization, underscoring the necessity to deeply investigate inter-architecture patterns with limited training samples.
In this study, we introduce a shared hypernetwork as an auxiliary task to dynamically generate weight for various architectures. This approach facilitates information exchange among diverse architectures through the soft weight-sharing mechanism of the hypernetwork, thereby enhancing the ability to capture architecture relationships. 
The auxiliary task is depicted by the green dashed box in Figure \ref{fig:HyperNAS_overview}.

Specifically, we employ a shared hypernetwork $H$ with parameters $\phi$ to generate weights for each operation within the architecture, taking as input a latent node feature $v$: 
\begin{equation}
    w_v=H(v;\phi|v\in \widetilde{\mathcal{V}})
\end{equation}
Notably, the node features $\widetilde{\mathcal{V}}$ are derived by the GCN encoder $G$,  which is shared with the neural predictor. The final architecture weights $w$ are obtained by applying the hypernetwork to all nodes. Formally, the forward process of the hypernetwork in our approach is expressed as:
\begin{equation}
\begin{split}
    w&=\{w^{(i)}\}_{i=1}^N\\
    &=\{w_v|v\in \widetilde{\mathcal{V}}^{(i)})\}_{i=1}^N\\
    &=\{H(v;\phi)|v\in G(z^{(i-1)}, \mathcal{A}^{(i)};\varphi)\}_{i=1}^N
\end{split}
\end{equation}
Here, $w^{(i)}$ denotes the set of generated weights for the $i$-th cell, and $\varphi$ represents the parameters of the GCN encoder.
Following common practice, we implement the hypernetwork $H$ as a multilayer perceptron (MLP). The hypernetwork maintains a fixed output dimension and incorporates multiple heads to accommodate operations with varying kernel sizes. Notably, $H$ is shared across all nodes.

The parameters generated by the hypernetwork are applied to the target network, enabling the computation of the loss $\mathcal{L}_{hyper}$ using an auxiliary dataset $\mathcal{D}_{aux}$. 
Notably, the hypernetwork does not have its own loss function but relies on the target network's specific task.  In this paper, we adopt the standard cross-entropy loss for $L_{hyper}$.

During training, we use the chain rule to compute the gradients of the loss with respect to both hypernetwork parameters $\phi$  and GCN parameters $\varphi$.  The gradient update rules for the hypernetwork and GCN can be formalized as follows:
\begin{equation}
    \nabla_{\phi} \mathcal{L}_{hyper}=\nabla_{w} \mathcal{L}_{hyper}\cdot\frac{\partial w}{\partial \phi}
\end{equation}
{
\small
\begin{equation}
    \nabla_{\varphi} \mathcal{L}_{hyper}=\{\nabla_{w_v} \mathcal{L}_{hyper}\cdot\frac{\partial w_v}{\partial v}\cdot\frac{\partial v}{\partial \varphi}|v\in G(z^{(i-1)},A^{(i)};\varphi)\}_{i=1}^N
\end{equation}
}
The training process jointly optimizes parameters of both $G$ and $H$ by propagating errors from the target network.

By integrating the hypernetwork as an auxiliary task, HyperNAS enhances the generalization capability of neural predictors through two primary mechanisms. First, the joint optimization of the hypernetwork and GCN encoder facilitates cross-architecture knowledge transfer and implicitly regularizes model complexity, thereby mitigating overfitting. Second, the dynamic parameter generation mechanism enables the model to adaptively adjust weight distributions based on the topological features of input architectures, thus expanding its coverage across the architectural space.

\subsection{Adaptive Multi-Task Objective}
Designing a multi-task loss function is crucial for balancing the trade-offs between different tasks, ensuring that each receives appropriate focus and resources. A common approach is Linear Scalarization~\cite{lin2019pareto}, which uses a linear weighted sum to combine the losses of all tasks. 
However, manually assigning task weights has significant drawbacks, such as extensive hyperparameter tuning, difficulty adapting to changing task relationships, and potential human bias leading to sub-optimal solutions.

To tackle these challenges, previous works~\cite{liebel2018auxiliary} have implemented a dynamic adaptive multi-task loss function that adjusts task weights in real-time based on uncertainties across different tasks, thereby improving balance and adaptability.
Namely, tasks with lower uncertainties (lower variances of predictions) are assigned with larger weights.
While such combination loss can converge towards a solution on the Pareto front, it disregards user preferences on the importance of various tasks, leading to limited adaptability and personalization.
To mitigate this limitation, we non-linearly scale task losses before computing their weighted sum, enabling the exploration of diverse solutions along the Pareto front.
The total loss $\mathcal{L}_{total}$ is defined as follows:
\begin{equation}
\mathcal{L}_{total}=\sum_{t\in\mathcal{T}}\frac{\mathcal{L}_t^{(q - 1)}}{2\cdot u_t^2}\cdot\mathcal{L}_t+ln(1+u_t^2).
\end{equation}
Here, $\mathcal{T}$ represents the set of all tasks, $\mathcal{L}_t$ is the loss for task $t$, $u_t$ is an adaptive weight associated with task $t$ that is a learnable parameter, and $q$ is the preference coefficient used to scale the task loss to obtain different models on the Pareto front. Moreover, we add the regularization term that prevents $u_t$ from becoming too large.

To understand the effect of the introduced hyperparameter $q$ in our loss function towards Pareto optimality, we start with the following definition:
\begin{definition}[Domination]
$(\phi^1, \varphi^1, \theta^1)$ are said to dominate $(\phi^2, \varphi^2, \theta^2)$ if and only if $\mathcal{L}_{t}(\phi^1, \varphi^1, \theta^1) \leq \mathcal{L}_{t}(\phi^2, \varphi^2, \theta^2)~\forall t \in \mathcal{T}$.
\end{definition}
\begin{definition}[Pareto optimality]
$(\phi^*, \varphi^*, \theta^*)$ is pareto optimal if no other $(\phi, \varphi, \theta)$ dominate $(\phi^*, \varphi^*, \theta^*)$.
\end{definition}
In other words, $(\phi^*, \varphi^*, \theta^*)$ is Pareto optimal if and only if any further update made on $(\phi^*, \varphi^*, \theta^*)$ will result in performance degradation for some tasks, \textit{i.e.}, $\sum_t^\mathcal{T} \alpha_t \nabla_{\phi^*, \varphi^*, \theta^*} \mathcal{L}_t(\phi^*, \varphi^*, \theta^*) = 0$,  where $\alpha_t$ is the task weight and $\sum_t^\mathcal{T} \alpha_t = 1.0$~\cite{lin2019pareto}.

In our case, it is evident that when the model converges, we have $\nabla_{\phi, \varphi, \theta} \mathcal{L}_{total} = \sum_{t\in\mathcal{T}}\frac{1}{2\cdot u_t^2}\cdot \nabla_{\phi, \varphi, \theta}(\mathcal{L}_t^q+ln(1+u_t^2)) = 0$.
Moreover, although it is not guaranteed that $\sum_{t\in\mathcal{T}}\frac{\mathcal{L}_t^{(q - 1)}}{2\cdot u_t^2} = 1.0$, our combination loss can be re-scaled to:
\begin{equation}
\mathcal{L}_{total}=\sum_{t\in\mathcal{T}}\frac{\mathcal{L}_t^{(q - 1)}}{2s\cdot u_t^2}\cdot s\mathcal{L}_t+ln(1+u_t^2),
\end{equation}
where $s$ is a scaler that satisfies $\sum_{t\in\mathcal{T}}\frac{\mathcal{L}_t^{(q - 1)}}{2s\cdot u_t^2} = 1.0$.
Thus, we show that our introduced preference coefficient $q$ only affects weights of different tasks without impacts on the Pareto optimality, and the obtained $(\phi, \varphi, \theta)$ by our adaptive multi-task objective loss is guaranteed to be Pareto optimal.

\section{Experiments}
\label{sec:experiments}
We evaluate the performance of HyperNAS across five different search spaces. First, we assess its ranking capability on NAS-Bench-101~\cite{ying2019bench} and NAS-Bench-201~\cite{dong2020bench} using various train-test splits. We then apply HyperNAS to search for CNN architectures on CIFAR-10 and ImageNet~\cite{krizhevsky2017imagenet} within the DARTS search space~\cite{liu2018darts}. In addition to standard benchmarks, we also explore architecture search based on MobileNet-V3~\cite{howard2019searching} and Vision Transformer (ViT)~\cite{dosovitskiy2020image}, which has emerged as a leading architecture in recent years.
To provide deeper insights into the effectiveness of HyperNAS, we conduct detailed ablation studies that highlight the contribution of each component. Due to space limitations, additional information, including full experimental settings, search results on the MobileNetV3, extended ablation analyses, and more, is provided in the supplementary material.

\subsection{Ranking Results on NAS Benchmarks}
\noindent\textbf{Implementation details.}

\begin{table}[t!]
\centering
\caption{Ranking results on NAS-Bench-101 and NAS-Bench-201. $^{\star}$: implemented with our settings. $^{\dag}$: results reported by PINAT~\cite{lu2023pinat}. EM: Embedding matrices. XFMR: Transformer.}
\tablestyle{6.3pt}{1}
\begin{tabular}{cc|ccccc|cccc}
\hline
\multirow{4}{*}{Method} & \multirow{4}{*}{Backbone} 
& \multicolumn{5}{c|}{NAS-Bench-101} 
& \multicolumn{4}{c}{NAS-Bench-201} \\
\cline{3-11}
&& 50 & 100 & 172 & 424 & 424 & 39 & 78 & 156 & 469 \\
&& (0.01\%) & (0.02\%) & (0.04\%) & (0.1\%) & (0.1\%) & (0.25\%) & (0.5\%) & (1\%) & (3\%) \\
&& \textit{test all} & \textit{test all} & \textit{test all} & \textit{test 100} & \textit{test all} & \textit{test all} & \textit{test all} & \textit{test all} & \textit{test all} \\
\hline\hline
SPOS$^\dag$ & Supernet & - & - & - & 0.193 & - & - & - & - & - \\
FairNAS$^\dag$ & Supernet & - & - & - & -0.232 & - & - & - & - & - \\
ReNAS$^\dag$ & LeNet-5  & - & - & - & 0.634 & 0.657 & - & - & - & - \\
NP$^\star$ & GCN & 0.662 & 0.639 & 0.624  & 0.685 & 0.693 & 0.655 & 0.626 & 0.696 & 0.757 \\
NAO$^\dag$ & LSTM  & - & 0.501 & 0.566 & 0.704 & 0.666 & - & 0.467 & 0.493 & 0.47 \\
GATES$^\dag$ & EM & - & 0.605 & 0.659 & 0.666 & 0.691 & - & - & - & - \\
Arch2Vec$^\dag$ & GIN  & - & 0.435 & 0.511 & 0.561 & 0.547 & - & 0.542 & 0.573 & 0.601 \\
D-VAE$^\dag$  & GNN  & - & 0.53 & 0.549 & 0.671 & 0.626 & - & - & - & - \\
GraphTrans$^\dag$ & XFMR & - & 0.33 & 0.472 & 0.600 & 0.602 & - & 0.409 & 0.55 & 0.594 \\
Graphormer$^\dag$ & XFMR & - & 0.564 & 0.580 & 0.596  & 0.611 & - & 0.505 & 0.63 & 0.68 \\
CTNAS$^\star$ & GCN & 0.431 & 0.601  & 0.628 & 0.751 & 0.720 & - & - & - & - \\
TNASP$^\dag$ & XFMR & - & 0.6  & 0.669 & 0.752 & 0.705 & - & 0.539 & 0.589 & 0.64 \\
GMAE-NAS$^\dag$ & GAT & - & 0.666 & 0.697 & 0.788 & 0.732 & - & - & - & - \\
PINAT$^\dag$ & XFMR & 0.580 & 0.679 & 0.715 & 0.801 & 0.772 & 0.494 & 0.549 & 0.631 & 0.706 \\
CAP$^\dag$ & GIN & - & 0.656 & 0.709 & 0.791 & 0.758 & - & 0.600 & 0.684 & 0.755 \\
\rowcolor[HTML]{EFF6FB} 
\textbf{HyperNAS} & \textbf{GCN} & \textbf{0.678} & \textbf{0.688} & \textbf{0.736} & \textbf{0.817}& \textbf{0.779} & \textbf{0.667} & \textbf{0.677} & \textbf{0.765} & \textbf{0.836} \\ \hline
\end{tabular}
\label{table:nasbench}
\vspace{-13pt}
\end{table}
In this study, we use a uniform discrete architecture encoding format~\cite{liu2022bridge}, where nodes represent operations and edges denote connections between them. 
To ensure complete architecture encodings, HyperNAS incorporates specific reduction operations, treating them as reduction cells. 
This approach results in a total of 11 cells per architecture in NAS-Bench-101 and 17 cells per architecture in NAS-Bench-201. 
Specifically, the 3rd and 7th cells are designated as reduction cells in NAS-Bench-101, and the 6th and 12th cells as reduction cells in NAS-Bench-201, with the remaining cells being normal cells. HyperNAS employs two separate GCNs to distinctly encode these two types of cells.  
We use 0.01\%, 0.02\%, 0.04\%, and 1\% of the data for NAS-Bench-101, and 0.25\%, 0.5\%, 1\%, and 3\% for NAS-Bench-201 as training sets. We evaluate the ranking abilities of neural predictors using Kendall’s Tau to measure the correlation between predicted and actual accuracy. 
\\
\noindent\textbf{Comparisons with SOTA methods.} 
Tables \ref{table:nasbench} present the ranking results on NAS-Bench-101 and NAS-Bench-201, respectively. 
HyperNAS achieves the highest scores across all data splits on both benchmarks. 
In particular, HyperNAS surpasses the current state-of-the-art PINAT~\cite{lu2023pinat} on the NAS-Bench-201 by notable margins of 0.173, 0.128, 0.134, and 0.130.
In conditions where training data is extremely deficient, such as 0.01\%  for NAS-Bench-101 and 0.25\% for NAS-Bench-201, HyperNAS achieves considerably higher Kendall’s Tau scores than competing baselines, underscoring its powerful generalization capabilities in few-shot training scenarios. 
\begin{wrapfigure}{r}{0.56\textwidth}
    \vspace{-3pt}
    \centering
    \captionsetup{type=table}
    \caption{Comparison results with zero-cost NAS methods. SP and KD denote Spearman and Kendall’s Tau correlation coefficients, respectively.}
    \label{table:zero-cost}
\tablestyle{2.3pt}{1.1}
\begin{tabular}{ccccccccc}
\hline
\multicolumn{1}{l}{\textbf{}} & \multicolumn{2}{c}{NB101-CF10} & \multicolumn{2}{c}{NB201-CF10} & \multicolumn{2}{c}{NB201-CF100} & \multicolumn{2}{c}{NB201-IMG16} \\ \cline{2-9} 
\multicolumn{1}{l}{\textbf{}} & SP & KD & SP & KD & SP & KD & SP & KD \\ \hline
Fisher & -0.28 & -0.20 & 0.50 & 0.37 & 0.54 & 0.40 & 0.48 & 0.36 \\
GradNorm & -0.25 & -0.17 & 0.58 & 0.42 & -0.63 & 0.47 & 0.57 & 0.42 \\
GraSP & 0.27 & 0.18 & 0.51 & 0.35 & 0.54 & 0.38 & 0.55 & 0.39 \\
L2Norm & 0.50 & 0.35 & 0.68 & 0.49 & 0.72 & 0.52 & 0.69 & 0.5 \\
SNIP & -0.19 & -0.14 & 0.58 & 0.43 & -0.63 & 0.47 & 0.57 & 0.42 \\
Synflow & 0.31 & 0.21 & 0.73 & 0.54 & 0.76 & 0.57 & 0.75 & 0.56 \\ \hline
NWOT & 0.31 & 0.21 & 0.77 & 0.58 & 0.80 & 0.62 & 0.77 & 0.59 \\
Zen & 0.59 & 0.42 & 0.35 & 0.27 & 0.35 & 0.28 & 0.39 & 0.29 \\
ZiCo & 0.63 & 0.46 & 0.74 & 0.54 & 0.78 & 0.58 & 0.79 & 0.60 \\
EZNAS & 0.68 & 0.45 & 0.83 & 0.65 & 0.82 & 0.65 & 0.78 & 0.61 \\
ParZC & 0.83 & 0.64 & 0.90 & 0.71 & 0.91 & 0.74 & 0.88 & 0.70 \\
\rowcolor[HTML]{EFF6FB} 
\textbf{HyperNAS} & \textbf{0.94} & \textbf{0.80} & \textbf{0.96} & \textbf{0.85} & \textbf{0.97} & \textbf{0.83} & \textbf{0.95} & \textbf{0.81} \\ \hline
\end{tabular}
\vspace{-15pt}
\end{wrapfigure}
\\
\noindent\textbf{Comparisons with zero-cost NAS methods.} Table \ref{table:zero-cost} summarizes the comparison results with recent popular zero-cost NAS methods~\cite{turner2019blockswap,abdelfattah2021zero,wang2020picking,lee2018snip,tanaka2020pruning,mellor2021neural,lin2021zen,li2023zico, dong2025parzc}. The methods are grouped into two categories: those based on pruning-inspired proxies and those derived from theoretical proxies. Following the same experimental setup as ParZC~\cite{dong2025parzc}, we find that HyperNAS consistently outperforms all existing approaches by a significant margin. These results confirm that HyperNAS achieves an excellent trade-off between computational efficiency and predictive accuracy.

\begin{table}[t!]
  \begin{minipage}{0.48\linewidth}
  \centering
  \caption{Performance comparison of architectures found by different NAS algorithms in DARTS search space on CIFAR-10. }
    \tablestyle{3pt}{1.1}
    \begin{tabular}{cccccc}
    \hline
    \textbf{Method} & \begin{tabular}[c]{@{}c@{}}Param\\ (M)\end{tabular} & \begin{tabular}[c]{@{}c@{}}Best acc\\ (\%)\end{tabular} & \begin{tabular}[c]{@{}c@{}}Average acc\\ (\%)\end{tabular} & \begin{tabular}[c]{@{}c@{}}Cost\\ (G$\cdot$D)\end{tabular} & \begin{tabular}[c]{@{}c@{}}Queries\\ (K)\end{tabular} \\ \hline
    NASNet-A & 3.3 & 97.35 & - & 1,800 & 20 \\
    AmoebaNet-A & 3.2 & - & 96.66±0.06 & 3,150 & 27 \\
    ENAS & 4.6 & 97.11 & - & 0.5 & - \\
    GHN & 5.7 & - & 97.16±0.07 & 0.8 & 1 \\
    DARTS & 3.3 & - & 97.24±0.09 & 4 & 19.5 \\ \hline
    PNAS & 3.2 & - & 97.17±0.07 & - & 12.8 \\
    NAO & 10.6 & 97.52 & - & 200 & 1 \\
    D-VAE & - & 94.8 & - & - & - \\
    GATES & 4.1 & 97.42 & - & - &  0.8\\
    Arch2vec-BO & 3.6 & 97.52 & 97.44±0.05 & 100 &  -\\
    GP-NAS & 3.9 & 96.21 & - & 0.9 & - \\
    BONAS-D & 3.3 & 97.57 & - & 10 & 4.8 \\
    BANANAS & 3.6 & - & 97.33±0.07 & 11.8 & - \\
    CATE & 3.5 & - & 97.45±0.08 & 3.3 & 0.3 \\
    CTNAS & 3.6 & - & 97.41±0.04 & 0.3 & - \\
    TNASP & 3.6 & 97.48 & 97.43±0.04 & 0.3 & 1 \\
    PINAT & 3.6 & 97.58 & 97.46±0.08 & 0.3 & 1 \\ 
    CAP & 3.3 & 97.58 & 97.46±0.09 & 3.3 &- \\ 
    \rowcolor[HTML]{EFF6FB} 
    \textbf{HyperNAS-P} & 3.8 & \textbf{97.61} & \textbf{97.55±0.00} & \textbf{0.1} & 1 \\
    \rowcolor[HTML]{EFF6FB} 
    \textbf{HyperNAS} & 4.8 & \textbf{97.60} & \textbf{97.48±0.01} & \textbf{0.3} & \textbf{0.2} \\ \hline
    \end{tabular}
    \label{table:darts}
    \end{minipage}
    \hspace{3mm}
    \begin{minipage}{0.48\linewidth}
   \centering
    \caption{Performance comparison of architectures searched by SOTA methods on ImageNet in the DARTS and ViT search spaces.}
    \label{table:darts_and_vit}
    \tablestyle{5pt}{0.965}
    \begin{tabular}{cccccc}
        \hline
        Method & 
        \begin{tabular}{@{}c@{}}Param\\(M)\end{tabular} & 
        \begin{tabular}{@{}c@{}}FLOPs\\(M/G)\end{tabular} & 
        \begin{tabular}{@{}c@{}}Top-1\\(\%)\end{tabular} & 
        \begin{tabular}{@{}c@{}}Top-5\\(\%)\end{tabular} & 
        \begin{tabular}{@{}c@{}}Queries\\(K)\end{tabular} \\ \hline
        \multicolumn{6}{c}{DARTS} \\ \hline
        PNAS & 5.1 & 588 & 74.2 & 91.9 & - \\
        NAO & 11.4 & 584 & 74.3 & 91.8 & 1 \\
        Arch2vec-B0 & 5.2 & 580 & 74.5 & - & - \\
        BONAS-D & 4.8 & 532 & 74.6 & 92.0 & 4.8 \\
        BANANAS & 5.1 & - & 73.7 & - & - \\
        DARTS & 4.9 & 595 & 73.1 & 91.0 & 19.5 \\
        GHN & 6.1 & 569 & 73.0 & 91.3 & 1 \\
        CATE & 5.0 & - & 73.9 & - & - \\
        PINAT & 5.2 & 583 & 75.1 & 92.5 & 1 \\ 
        \rowcolor[HTML]{EFF6FB} 
        \textbf{HyperNAS} & 6.6 & 743 & \textbf{75.4} & \textbf{92.5} & \textbf{0.2} \\ \hline
        \multicolumn{6}{c}{ViT-Small} \\ \hline
        DeiT-S & 22.1 & 4.7G & 79.9 & 95.00 & - \\
        ViT-S/16 & 22.1 & 4.7G & 78.8 & - & - \\
        NASViT-A0 & - & - & 78.2 & - & 10 \\
        AutoFormer & 22.9 & 5.1G & 81.7 & 95.7 & 1 \\
        \rowcolor[HTML]{EFF6FB} 
        \textbf{HyperNAS} & \textbf{22.8} & \textbf{4.8G} & \textbf{81.8} & \textbf{95.7} & \textbf{0.2} \\ \hline
        \multicolumn{6}{c}{ViT-Base} \\ \hline
        ViT-B/16 & 86 & 18G & 79.7 & - & - \\
        Deit-B & 86 & 18G & 81.8 & 95.6 & - \\
        NASViT-A3 & - & - & 81.0 & - & 10 \\
        AutoFormer & 54 & 11G & 82.4 & 95.7 & 1 \\
        \rowcolor[HTML]{EFF6FB} 
        \textbf{HyperNAS} & \textbf{54} & \textbf{11G} & \textbf{82.4} & \textbf{95.8} & \textbf{0.2} \\ \hline
    \end{tabular}
\label{table:imagenet}
\end{minipage}
\vspace{-10pt}
\end{table}

\subsection{Search Results on CIFAR-10}

\noindent\textbf{Implementation details.} 
We utilize two distinct GCNs to encode different types of cells. To validate the robustness of our method, we implement two paradigms: HyperNAS-P and HyperNAS. HyperNAS-P is a single-task approach that focuses solely on the neural predictor with the global encoding scheme, excluding the auxiliary task. In line with PINAT~\cite{lu2023pinat}, HyperNAS-P is trained using 1000 randomly sampled architecture-accuracy pairs, whereas HyperNAS employs only 200 sampled pairs. 
We use these well-trained predictors for performance evaluation and implement an evolutionary algorithm as the search strategy, adhering to the same settings as PINAT~\cite{lu2023pinat}.
The searched architectures are retrained using the DARTS~\cite{liu2018darts} settings. 
To minimize randomness, we retrain the top five searched architectures and report the average metrics across three random seeds.

\noindent\textbf{Comparisons with SOTA methods.} 
Table \ref{table:darts} summarizes the evaluation results of the architectures searched within the DARTS search space on CIFAR-10. The first row shows the performance of well-known NAS methods, while the second row details the predictor-based NAS methods. Among baselines with a similar single-task paradigm, HyperNAS-P achieves the highest accuracy and robustness with the least search cost of 0.1 GPU days, demonstrating the effectiveness and efficiency of our global encoding scheme. In contrast, HyperNAS as a dual-task paradigm outperforms other competing methods with just 200 queries, underscoring its great advantages in few-shot scenarios and the benefits of introducing an auxiliary supervised task. The inclusion of the hypernetwork slightly raises the search cost for HyperNAS, but this increase is negligible compared to the prohibitive expense of acquiring large architecture-accuracy pairs.

\subsection{Search Results on ImageNet}
\textcolor{black}{\noindent\textbf{Implementation details.} }
To evaluate the robustness of HyperNAS, we first transfer the architecture discovered on CIFAR-10 within the DARTS search space to ImageNet, adhering to the same transfer configuration as DARTS~\cite{liu2018darts} for retraining.
Subsequently, we conduct an architecture search in the ViT search space using ImageNet. We gather architecture-accuracy pairs from a well-trained supernet of Autoformer~\cite{chen2021autoformer} and follow identical evolutionary settings. We limit the parameter count to under 23M for ViT-Small and 55M for ViT-Base. The final searched architecture is evaluated by fine-tuning for 30 epochs with a learning rate of 1e-5. The search results on MobileNetV3 are presented in Appendix \ref{sec:mobilenet}.
\begin{wrapfigure}{R}{0.5\linewidth}
   \vspace{-20pt}
    \centering
    \captionsetup[subfigure]{margin=2pt}
    \subfloat[Ablation studies on ranking abilities of the neural predictor.]{
        \label{subfig:pred_101}
        \includegraphics[width=0.47\linewidth]{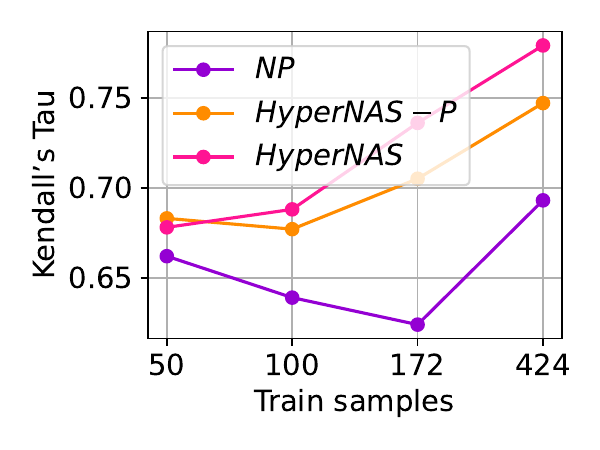}
    }
    \subfloat[Ablation studies on parameter prediction of the hypernetwork.]{
        \label{subfig:hyper_101} 
        \includegraphics[width=0.50\linewidth]{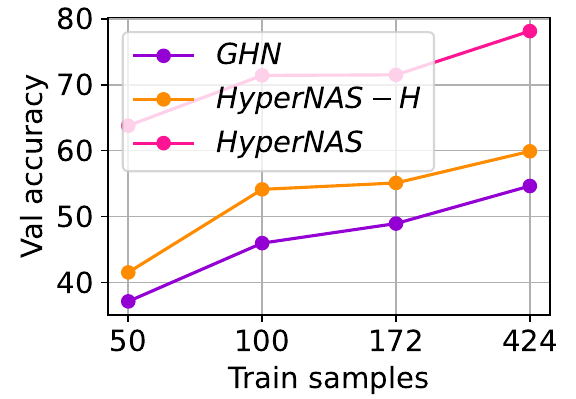}
    }
    \caption{Ablation studies on NAS-Bench-101.}
    \label{fig:ablate_predictor}
    \vspace{-10pt}
\end{wrapfigure}


\textcolor{black}{\noindent\textbf{Comparisons with SOTA methods.} }
Table \ref{table:imagenet} presents the evaluation results on ImageNet. In the DARTS search space, HyperNAS achieves a top-1 transfer accuracy of 75.4\%, showcasing the superior generalization ability of its discovered architectures. For ViT, HyperNAS outperforms manually designed models such as DeiT~\cite{touvron2021training} and ViT~\cite{dosovitskiy2020image}, while delivering comparable performance to automated methods like AutoFormer~\cite{chen2021autoformer} with only 20\% of the query cost. Importantly, HyperNAS completes the search in under \textbf{1 minute} via performance prediction, whereas AutoFormer requires approximately 1 GPU day due to test-set inference at each evaluation step. This highlights the significant efficiency advantage of HyperNAS without compromising on performance.


\subsection{Ablations and Hyperparameter Analysis}
\begin{figure}[t!] 
\vspace{-10pt}
    \centering
    \captionsetup[subfigure]{margin=2pt}
 \subfloat[NP]{
        \label{subfig:NP_feat}
        \centering
\includegraphics[width=0.24\linewidth]
{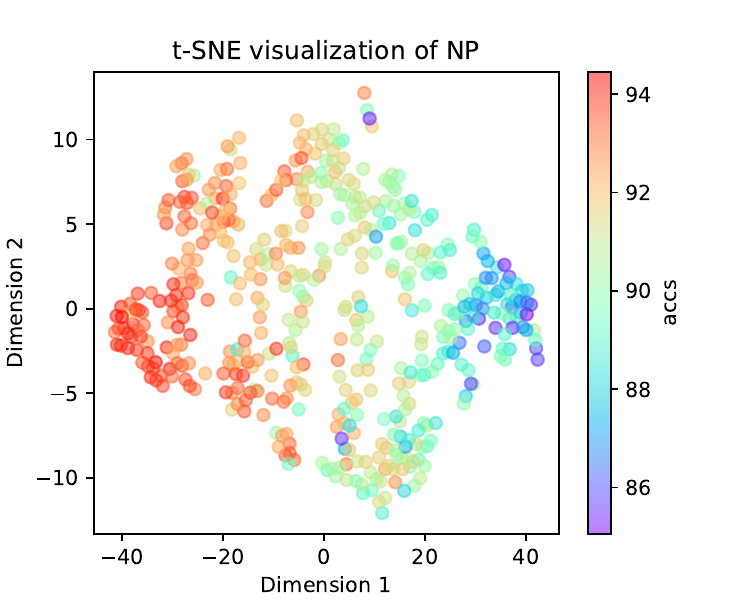}}
    \subfloat[HyperNAS-P+cellFeat]{
        \label{subfig:HyperNAS-P_cellFeat}
        \centering
\includegraphics[width=0.24\linewidth]{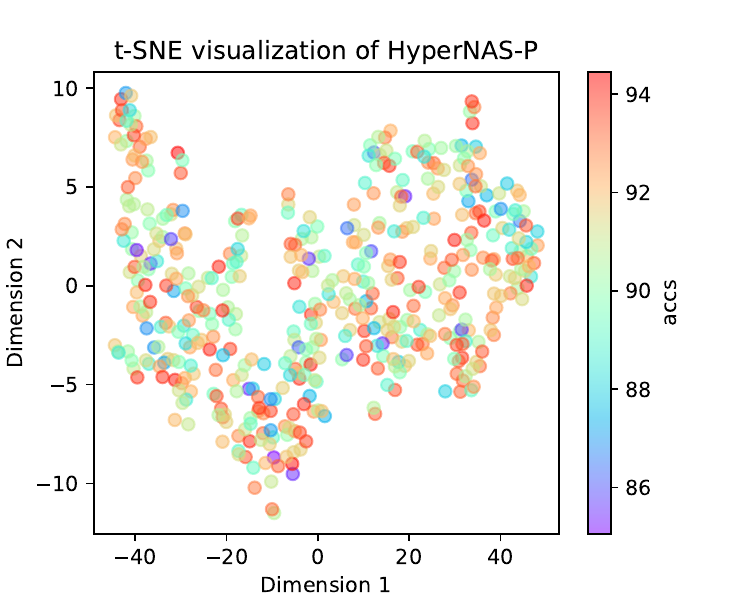}}
    \subfloat[HyperNAS-P + cellPos]{
        \label{subfig:HyperNAS-P_cellPos}
        \centering   
         \includegraphics[width=0.24\linewidth]{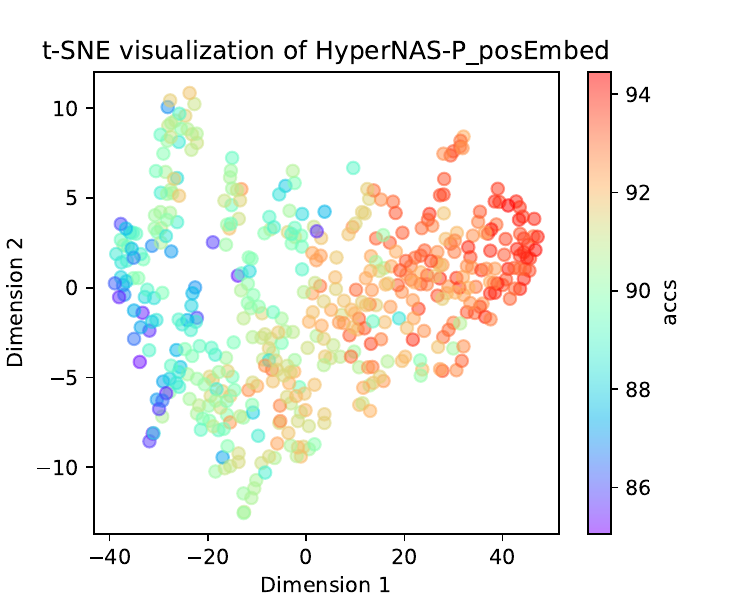}
        }
    \subfloat[HyperNAS + cellPos]{
        \label{subfig:HyperNAS_cellPos}
        \centering     
        \includegraphics[width=0.24\linewidth]{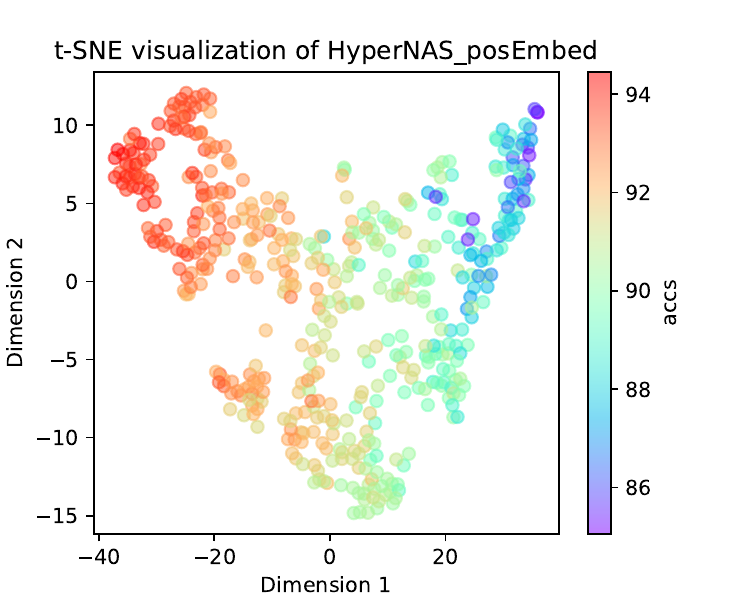}
        }
    \caption{t-SNE visualization of learned architecture features. }
    \label{fig:arch_feat_visualize}
    \vspace{-15pt}
\end{figure}

\noindent\textbf{Effect of the global encoding scheme.} 
To verify the effect of the proposed global encoding scheme, we implement two methods with identical GCN encoders: NP and HyperNAS-P. \begin{wrapfigure}{r}{0.3\linewidth}
  \vspace{-5pt}
  \centering
  \includegraphics[width=\linewidth]{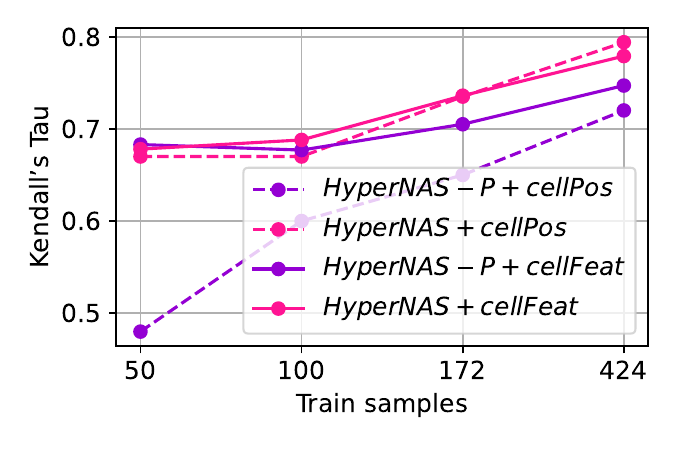}
  \caption{Ranking results of neural predictors with different cell connectivity strategies}
  \label{fig:cell_connectivity}
  \vspace{-10pt}
\end{wrapfigure}
NP utilizes the traditional cell encoding scheme, while HyperNAS-P employs our proposed global encoding scheme. As illustrated in Figure \ref{subfig:pred_101}, HyperNAS-P outperforms NP with notable margins across all data splits. 
This substantial improvement demonstrates that our global encoding scheme effectively boosts architecture representations, leading to more accurate performance predictions.
For deeper insights, we randomly visualized 500 sampled architectures from the search space using the t-SNE tool, as depicted in Figures \ref{subfig:NP_feat} and \ref{subfig:HyperNAS-P_cellFeat}. 
Surprisingly, unlike the NP method, the architecture features derived from HyperNAS-P exhibit no correlations with accuracy.
We attribute this phenomenon to the cell connectivity strategy within the global encoding scheme, specifically cell features that encapsulate context information from preceding cells. 
This hypothesis finds validation in Figure \ref{subfig:HyperNAS-P_cellPos}, where cell features were replaced with cell position embeddings to represent inter-cell dependencies.
Moreover, the superior ranking performance of our cell connectivity strategy, as evidenced in Figure \ref{fig:cell_connectivity}, underscores the pivotal role of cell features in enhancing architecture representations.

\noindent\textbf{Effect of the hypernetwork.} 
We comprehensively examined the impact of hypernetwork from multiple perspectives. Firstly, Figure \ref{subfig:pred_101} illustrates that HyperNAS outperforms HyperNAS-P in ranking performance across most data splits, highlighting the hypernetwork utilizing soft wight-sharing mechanism effectively boosts information exchange among various architectures.
Secondly, Figures  \ref{subfig:HyperNAS-P_cellPos} and \ref{subfig:HyperNAS_cellPos} reveal that the inclusion of the hypernetwork significantly improves the clustering of architectural features, validating its powerful abilities in learning underlying architecture patterns.
Lastly, Figure \ref{fig:cell_connectivity} shows that HyperNAS-P performs poorly with cell position embeddings, but the inclusion of the hypernetwork effectively bridges this performance gap. 
This evidence underscores the critical role of the hypernetwork as an auxiliary task, significantly enhancing architecture representations.

\noindent\textbf{Effect of the multi-task paradigm.}
To investigate the trade-off in our multi-task paradigm, we conduct experiments on HyperNAS, HyperNAS-P, and HyperNAS-H, where HyperNAS-H excludes the neural predictor and focuses sorely on hypernetwork. As shown in Figures \ref{subfig:pred_101} and \ref{subfig:hyper_101}, the predictor from HyperNAS shows superior ranking abilities compared to HyperNAS-P, while the hypernetwork from HyperNAS achieves higher validation accuracy than HyperNAS-H. These findings demonstrate that our multi-task paradigm improves the performance of both tasks, proving the compatibility of these two tasks and the effectiveness of our multi-task loss function.


\vspace{-5pt}
\section{Conclusion}
\label{sec:conclusion}
\vspace{-5pt}
This paper introduces HyperNAS, a novel neural predictor paradigm aimed at enhancing architecture representation learning. HyperNAS introduces an auxiliary task to provide additional supervision and regularization, which encourages the model to learn more generalizable features rather than overfitting to training samples. Extensive experiments demonstrate the superior robustness and efficiency of HyperNAS.\\
\textbf{Future work.} Although HyperNAS significantly reduces the need for extensive training samples, its time cost may still be computationally demanding for extremely large architectures due to the hypernetwork. In future work, we plan to retain the auxiliary task paradigm while focusing on optimizing efficiency to further enhance scalability.



\newpage
\bibliographystyle{unsrt}
\bibliography{ref}

\begin{thebibliography}{10}

\bibitem{howard2019searching}
Andrew Howard, Mark Sandler, Grace Chu, Liang-Chieh Chen, Bo~Chen, Mingxing Tan, Weijun Wang, Yukun Zhu, Ruoming Pang, Vijay Vasudevan, et~al.
\newblock Searching for mobilenetv3.
\newblock In {\em Proceedings of the IEEE/CVF international conference on computer vision}, pages 1314--1324, 2019.

\bibitem{wang2020fcos}
Ning Wang, Yang Gao, Hao Chen, Peng Wang, Zhi Tian, Chunhua Shen, and Yanning Zhang.
\newblock Nas-fcos: Fast neural architecture search for object detection.
\newblock In {\em proceedings of the IEEE/CVF conference on computer vision and pattern recognition}, pages 11943--11951, 2020.

\bibitem{sarah2024llama}
Anthony Sarah, Sharath~Nittur Sridhar, Maciej Szankin, and Sairam Sundaresan.
\newblock Llama-nas: Efficient neural architecture search for large language models.
\newblock {\em arXiv preprint arXiv:2405.18377}, 2024.

\bibitem{chen2021neural}
Wuyang Chen, Xinyu Gong, and Zhangyang Wang.
\newblock Neural architecture search on imagenet in four gpu hours: A theoretically inspired perspective.
\newblock {\em arXiv preprint arXiv:2102.11535}, 2021.

\bibitem{shen2022efficient}
Junhong Shen, Misha Khodak, and Ameet Talwalkar.
\newblock Efficient architecture search for diverse tasks.
\newblock {\em Advances in Neural Information Processing Systems}, 35:16151--16164, 2022.

\bibitem{cai2018proxylessnas}
Han Cai, Ligeng Zhu, and Song Han.
\newblock Proxylessnas: Direct neural architecture search on target task and hardware.
\newblock In {\em International Conference on Learning Representations}, 2018.

\bibitem{zoph2016neural}
Barret Zoph and Quoc Le.
\newblock Neural architecture search with reinforcement learning.
\newblock In {\em International Conference on Learning Representations}, 2016.

\bibitem{real2019regularized}
Esteban Real, Alok Aggarwal, Yanping Huang, and Quoc~V Le.
\newblock Regularized evolution for image classifier architecture search.
\newblock In {\em Proceedings of the AAAI Conference on Artificial Intelligence}, volume~33, pages 4780--4789, 2019.

\bibitem{white2023neural}
Colin White, Mahmoud Safari, Rhea Sukthanker, Binxin Ru, Thomas Elsken, Arber Zela, Debadeepta Dey, and Frank Hutter.
\newblock Neural architecture search: Insights from 1000 papers.
\newblock {\em arXiv preprint arXiv:2301.08727}, 2023.

\bibitem{liu2018darts}
Hanxiao Liu, Karen Simonyan, and Yiming Yang.
\newblock Darts: Differentiable architecture search.
\newblock In {\em International Conference on Learning Representations}, 2018.

\bibitem{chu2020fair}
Xiangxiang Chu, Tianbao Zhou, Bo~Zhang, and Jixiang Li.
\newblock Fair darts: Eliminating unfair advantages in differentiable architecture search.
\newblock In {\em European conference on computer vision}, pages 465--480. Springer, 2020.

\bibitem{luo2018neural}
Renqian Luo, Fei Tian, Tao Qin, Enhong Chen, and Tie-Yan Liu.
\newblock Neural architecture optimization.
\newblock {\em Advances in neural information processing systems}, 31, 2018.

\bibitem{wen2020neural}
Wei Wen, Hanxiao Liu, Yiran Chen, Hai Li, Gabriel Bender, and Pieter-Jan Kindermans.
\newblock Neural predictor for neural architecture search.
\newblock In {\em European conference on computer vision}, pages 660--676. Springer, 2020.

\bibitem{chen2021contrastive}
Yaofo Chen, Yong Guo, Qi~Chen, Minli Li, Wei Zeng, Yaowei Wang, and Mingkui Tan.
\newblock Contrastive neural architecture search with neural architecture comparators.
\newblock In {\em Proceedings of the IEEE/CVF conference on computer vision and pattern recognition}, pages 9502--9511, 2021.

\bibitem{lu2023pinat}
Shun Lu, Yu~Hu, Peihao Wang, Yan Han, Jianchao Tan, Jixiang Li, Sen Yang, and Ji~Liu.
\newblock Pinat: A permutation invariance augmented transformer for nas predictor.
\newblock In {\em Proceedings of the AAAI Conference on Artificial Intelligence}, volume~37, pages 8957--8965, 2023.

\bibitem{yi2023nar}
Yun Yi, Haokui Zhang, Wenze Hu, Nannan Wang, and Xiaoyu Wang.
\newblock Nar-former: Neural architecture representation learning towards holistic attributes prediction.
\newblock In {\em Proceedings of the IEEE/CVF Conference on Computer Vision and Pattern Recognition}, pages 7715--7724, 2023.

\bibitem{ji2024cap}
Han Ji, Yuqi Feng, and Yanan Sun.
\newblock Cap: A context-aware neural predictor for nas.
\newblock {\em arXiv preprint arXiv:2406.02056}, 2024.

\bibitem{xie2021weight}
Lingxi Xie, Xin Chen, Kaifeng Bi, Longhui Wei, Yuhui Xu, Lanfei Wang, Zhengsu Chen, An~Xiao, Jianlong Chang, Xiaopeng Zhang, et~al.
\newblock Weight-sharing neural architecture search: A battle to shrink the optimization gap.
\newblock {\em ACM Computing Surveys (CSUR)}, 54(9):1--37, 2021.

\bibitem{song2024efficient}
Xiaotian Song, Xiangning Xie, Zeqiong Lv, Gary~G Yen, Weiping Ding, Jiancheng Lv, and Yanan Sun.
\newblock Efficient evaluation methods for neural architecture search: A survey.
\newblock {\em IEEE Transactions on Artificial Intelligence}, 2024.

\bibitem{akhauri2024encodings}
Yash Akhauri and Mohamed~S Abdelfattah.
\newblock Encodings for prediction-based neural architecture search.
\newblock {\em arXiv preprint arXiv:2403.02484}, 2024.

\bibitem{yan2021cate}
Shen Yan, Kaiqiang Song, Fei Liu, and Mi~Zhang.
\newblock Cate: Computation-aware neural architecture encoding with transformers.
\newblock In {\em International Conference on Machine Learning}, pages 11670--11681. PMLR, 2021.

\bibitem{yan2020does}
Shen Yan, Yu~Zheng, Wei Ao, Xiao Zeng, and Mi~Zhang.
\newblock Does unsupervised architecture representation learning help neural architecture search?
\newblock {\em Advances in neural information processing systems}, 33:12486--12498, 2020.

\bibitem{pham2018efficient}
Hieu Pham, Melody Guan, Barret Zoph, Quoc Le, and Jeff Dean.
\newblock Efficient neural architecture search via parameters sharing.
\newblock In {\em International conference on machine learning}, pages 4095--4104. PMLR, 2018.

\bibitem{ying2019bench}
Chris Ying, Aaron Klein, Eric Christiansen, Esteban Real, Kevin Murphy, and Frank Hutter.
\newblock Nas-bench-101: Towards reproducible neural architecture search.
\newblock In {\em International conference on machine learning}, pages 7105--7114. PMLR, 2019.

\bibitem{dong2020bench}
Xuanyi Dong and Yi~Yang.
\newblock Nas-bench-201: Extending the scope of reproducible neural architecture search.
\newblock {\em arXiv preprint arXiv:2001.00326}, 2020.

\bibitem{jing2022graph}
Kun Jing, Jungang Xu, and Pengfei Li.
\newblock Graph masked autoencoder enhanced predictor for neural architecture search.
\newblock In {\em IJCAI}, pages 3114--3120, 2022.

\bibitem{chauhan2023brief}
Vinod~Kumar Chauhan, Jiandong Zhou, Ping Lu, Soheila Molaei, and David~A Clifton.
\newblock A brief review of hypernetworks in deep learning.
\newblock {\em arXiv preprint arXiv:2306.06955}, 2023.

\bibitem{liu2021homogeneous}
Yuqiao Liu, Yehui Tang, and Yanan Sun.
\newblock Homogeneous architecture augmentation for neural predictor.
\newblock In {\em Proceedings of the IEEE/CVF International Conference on Computer Vision}, pages 12249--12258, 2021.

\bibitem{xie2023architecture}
Xiangning Xie, Yanan Sun, Yuqiao Liu, Mengjie Zhang, and Kay~Chen Tan.
\newblock Architecture augmentation for performance predictor via graph isomorphism.
\newblock {\em IEEE Transactions on Cybernetics}, 54(3):1828--1840, 2023.

\bibitem{lyu2021multiobjective}
Bo~Lyu, Shiping Wen, Kaibo Shi, and Tingwen Huang.
\newblock Multiobjective reinforcement learning-based neural architecture search for efficient portrait parsing.
\newblock {\em IEEE Transactions on Cybernetics}, 53(2):1158--1169, 2021.

\bibitem{movahedi2022lambda}
Sajad Movahedi, Melika Adabinejad, Ayyoob Imani, Arezou Keshavarz, Mostafa Dehghani, Azadeh Shakery, and Babak~N Araabi.
\newblock $\backslash lambda $-darts: Mitigating performance collapse by harmonizing operation selection among cells.
\newblock {\em arXiv preprint arXiv:2210.07998}, 2022.

\bibitem{zhang2023shiftnas}
Mingyang Zhang, Xinyi Yu, Haodong Zhao, and Linlin Ou.
\newblock Shiftnas: Improving one-shot nas via probability shift.
\newblock In {\em Proceedings of the IEEE/CVF International Conference on Computer Vision}, pages 5919--5928, 2023.

\bibitem{liu2022bridge}
Yuqiao Liu, Yehui Tang, Zeqiong Lv, Yunhe Wang, and Yanan Sun.
\newblock Bridge the gap between architecture spaces via a cross-domain predictor.
\newblock {\em Advances in Neural Information Processing Systems}, 35:13355--13366, 2022.

\bibitem{tang2020semi}
Yehui Tang, Yunhe Wang, Yixing Xu, Hanting Chen, Boxin Shi, Chao Xu, Chunjing Xu, Qi~Tian, and Chang Xu.
\newblock A semi-supervised assessor of neural architectures.
\newblock In {\em proceedings of the IEEE/CVF conference on computer vision and pattern recognition}, pages 1810--1819, 2020.

\bibitem{liu2018progressive}
Chenxi Liu, Barret Zoph, Maxim Neumann, Jonathon Shlens, Wei Hua, Li-Jia Li, Li~Fei-Fei, Alan Yuille, Jonathan Huang, and Kevin Murphy.
\newblock Progressive neural architecture search.
\newblock In {\em Proceedings of the European conference on computer vision (ECCV)}, pages 19--34, 2018.

\bibitem{white2021bananas}
Colin White, Willie Neiswanger, and Yash Savani.
\newblock Bananas: Bayesian optimization with neural architectures for neural architecture search.
\newblock In {\em Proceedings of the AAAI conference on artificial intelligence}, volume~35, pages 10293--10301, 2021.

\bibitem{xu2021renas}
Yixing Xu, Yunhe Wang, Kai Han, Yehui Tang, Shangling Jui, Chunjing Xu, and Chang Xu.
\newblock Renas: Relativistic evaluation of neural architecture search.
\newblock In {\em Proceedings of the IEEE/CVF conference on computer vision and pattern recognition}, pages 4411--4420, 2021.

\bibitem{li2020neural}
Wei Li, Shaogang Gong, and Xiatian Zhu.
\newblock Neural graph embedding for neural architecture search.
\newblock In {\em Proceedings of the AAAI Conference on Artificial Intelligence}, volume~34, pages 4707--4714, 2020.

\bibitem{shi2020bridging}
Han Shi, Renjie Pi, Hang Xu, Zhenguo Li, James Kwok, and Tong Zhang.
\newblock Bridging the gap between sample-based and one-shot neural architecture search with bonas.
\newblock {\em Advances in Neural Information Processing Systems}, 33:1808--1819, 2020.

\bibitem{lu2021tnasp}
Shun Lu, Jixiang Li, Jianchao Tan, Sen Yang, and Ji~Liu.
\newblock Tnasp: A transformer-based nas predictor with a self-evolution framework.
\newblock {\em Advances in Neural Information Processing Systems}, 34:15125--15137, 2021.

\bibitem{ha2017hypernetworks}
D~Ha, AM~Dai, and QV~Le.
\newblock Hypernetworks. international conference on learning representations, 2017.

\bibitem{zhao2020meta}
Dominic Zhao, Seijin Kobayashi, Jo{\~a}o Sacramento, and Johannes von Oswald.
\newblock Meta-learning via hypernetworks.
\newblock In {\em 4th Workshop on Meta-Learning at NeurIPS 2020 (MetaLearn 2020)}. NeurIPS, 2020.

\bibitem{beck2023hypernetworks}
Jacob Beck, Matthew~Thomas Jackson, Risto Vuorio, and Shimon Whiteson.
\newblock Hypernetworks in meta-reinforcement learning.
\newblock In {\em Conference on Robot Learning}, pages 1478--1487. PMLR, 2023.

\bibitem{cho2024hypernetwork}
Woojin Cho, Kookjin Lee, Donsub Rim, and Noseong Park.
\newblock Hypernetwork-based meta-learning for low-rank physics-informed neural networks.
\newblock {\em Advances in Neural Information Processing Systems}, 36, 2024.

\bibitem{von2019continual}
Johannes Von~Oswald, Christian Henning, Benjamin~F Grewe, and Jo{\~a}o Sacramento.
\newblock Continual learning with hypernetworks.
\newblock {\em arXiv preprint arXiv:1906.00695}, 2019.

\bibitem{chandra2023continual}
Dupati~Srikar Chandra, Sakshi Varshney, PK~Srijith, and Sunil Gupta.
\newblock Continual learning with dependency preserving hypernetworks.
\newblock In {\em Proceedings of the IEEE/CVF Winter Conference on Applications of Computer Vision}, pages 2339--2348, 2023.

\bibitem{hemati2023partial}
Hamed Hemati, Vincenzo Lomonaco, Davide Bacciu, and Damian Borth.
\newblock Partial hypernetworks for continual learning.
\newblock In {\em Conference on Lifelong Learning Agents}, pages 318--336. PMLR, 2023.

\bibitem{ratzlaff2019hypergan}
Neale Ratzlaff and Li~Fuxin.
\newblock Hypergan: A generative model for diverse, performant neural networks.
\newblock In {\em International Conference on Machine Learning}, pages 5361--5369. PMLR, 2019.

\bibitem{schurholt2022hyper}
Konstantin Sch{\"u}rholt, Boris Knyazev, Xavier Gir{\'o}-i Nieto, and Damian Borth.
\newblock Hyper-representations as generative models: Sampling unseen neural network weights.
\newblock {\em Advances in Neural Information Processing Systems}, 35:27906--27920, 2022.

\bibitem{do2020structural}
Manh~Tuan Do, Se-eun Yoon, Bryan Hooi, and Kijung Shin.
\newblock Structural patterns and generative models of real-world hypergraphs.
\newblock In {\em Proceedings of the 26th ACM SIGKDD international conference on knowledge discovery \& data mining}, pages 176--186, 2020.

\bibitem{brock2017smash}
Andrew Brock, Theodore Lim, James~M Ritchie, and Nick Weston.
\newblock Smash: one-shot model architecture search through hypernetworks.
\newblock {\em arXiv preprint arXiv:1708.05344}, 2017.

\bibitem{zhang2018graph}
Chris Zhang, Mengye Ren, and Raquel Urtasun.
\newblock Graph hypernetworks for neural architecture search.
\newblock {\em arXiv preprint arXiv:1810.05749}, 2018.

\bibitem{lin2019pareto}
Xi~Lin, Hui-Ling Zhen, Zhenhua Li, Qing-Fu Zhang, and Sam Kwong.
\newblock Pareto multi-task learning.
\newblock {\em Advances in neural information processing systems}, 32, 2019.

\bibitem{liebel2018auxiliary}
Lukas Liebel and Marco K{\"o}rner.
\newblock Auxiliary tasks in multi-task learning.
\newblock {\em arXiv preprint arXiv:1805.06334}, 2018.

\bibitem{krizhevsky2017imagenet}
Alex Krizhevsky, Ilya Sutskever, and Geoffrey~E Hinton.
\newblock Imagenet classification with deep convolutional neural networks.
\newblock {\em Communications of the ACM}, 60(6):84--90, 2017.

\bibitem{dosovitskiy2020image}
Alexey Dosovitskiy, Lucas Beyer, Alexander Kolesnikov, Dirk Weissenborn, Xiaohua Zhai, Thomas Unterthiner, Mostafa Dehghani, Matthias Minderer, Georg Heigold, Sylvain Gelly, et~al.
\newblock An image is worth 16x16 words: Transformers for image recognition at scale.
\newblock {\em arXiv preprint arXiv:2010.11929}, 2020.

\bibitem{turner2019blockswap}
Jack Turner, Elliot~J Crowley, Michael O'Boyle, Amos Storkey, and Gavin Gray.
\newblock Blockswap: Fisher-guided block substitution for network compression on a budget.
\newblock {\em arXiv preprint arXiv:1906.04113}, 2019.

\bibitem{abdelfattah2021zero}
Mohamed~S Abdelfattah, Abhinav Mehrotra, {\L}ukasz Dudziak, and Nicholas~D Lane.
\newblock Zero-cost proxies for lightweight nas.
\newblock {\em arXiv preprint arXiv:2101.08134}, 2021.

\bibitem{wang2020picking}
Chaoqi Wang, Guodong Zhang, and Roger Grosse.
\newblock Picking winning tickets before training by preserving gradient flow.
\newblock {\em arXiv preprint arXiv:2002.07376}, 2020.

\bibitem{lee2018snip}
Namhoon Lee, Thalaiyasingam Ajanthan, and Philip~HS Torr.
\newblock Snip: Single-shot network pruning based on connection sensitivity.
\newblock {\em arXiv preprint arXiv:1810.02340}, 2018.

\bibitem{tanaka2020pruning}
Hidenori Tanaka, Daniel Kunin, Daniel~L Yamins, and Surya Ganguli.
\newblock Pruning neural networks without any data by iteratively conserving synaptic flow.
\newblock {\em Advances in neural information processing systems}, 33:6377--6389, 2020.

\bibitem{mellor2021neural}
Joe Mellor, Jack Turner, Amos Storkey, and Elliot~J Crowley.
\newblock Neural architecture search without training.
\newblock In {\em International conference on machine learning}, pages 7588--7598. PMLR, 2021.

\bibitem{lin2021zen}
Ming Lin, Pichao Wang, Zhenhong Sun, Hesen Chen, Xiuyu Sun, Qi~Qian, Hao Li, and Rong Jin.
\newblock Zen-nas: A zero-shot nas for high-performance image recognition.
\newblock In {\em Proceedings of the IEEE/CVF International Conference on Computer Vision}, pages 347--356, 2021.

\bibitem{li2023zico}
Guihong Li, Yuedong Yang, Kartikeya Bhardwaj, and Radu Marculescu.
\newblock Zico: Zero-shot nas via inverse coefficient of variation on gradients.
\newblock {\em arXiv preprint arXiv:2301.11300}, 2023.

\bibitem{dong2025parzc}
Peijie Dong, Lujun Li, Zhenheng Tang, Xiang Liu, Zimian Wei, Qiang Wang, and Xiaowen Chu.
\newblock Parzc: Parametric zero-cost proxies for efficient nas.
\newblock In {\em Proceedings of the AAAI Conference on Artificial Intelligence}, volume~39, pages 16327--16335, 2025.

\bibitem{chen2021autoformer}
Minghao Chen, Houwen Peng, Jianlong Fu, and Haibin Ling.
\newblock Autoformer: Searching transformers for visual recognition.
\newblock In {\em Proceedings of the IEEE/CVF international conference on computer vision}, pages 12270--12280, 2021.

\bibitem{touvron2021training}
Hugo Touvron, Matthieu Cord, Matthijs Douze, Francisco Massa, Alexandre Sablayrolles, and Herv{\'e} J{\'e}gou.
\newblock Training data-efficient image transformers \& distillation through attention.
\newblock In {\em International conference on machine learning}, pages 10347--10357. PMLR, 2021.

\end{thebibliography}

\newpage
\appendix

\section{Implement Details}
\noindent\textbf{Training settings.} During each training step, we randomly select an architecture-accuracy pair and a batch of 128 images for training HyperNAS. 
\begin{wrapfigure}{r}{0.40\textwidth}
\vspace{-5pt}
    \centering
    \captionsetup{type=table}
    \caption{Predefined ViT search space}
    \label{table:vit-sp}
\tablestyle{5pt}{1.3}
\begin{tabular}{c|cc}
\hline
 & Supernet-small & Supernet-base \\ \hline
Embed Dim & (320, 448, 64) & (528, 624, 48) \\
Q-K-V Dim & (320, 448, 64) & (512, 640, 64) \\
MLP Ratio & (3, 4, 0.5) & (3, 4, 0.5) \\
Head Num & (5, 7, 1) & (8, 10, 1) \\
Depth Num & (12, 14, 1) & (14, 16, 1) \\ \hline
Params Range & 14–34M & 42–75M \\ \hline
\end{tabular}
\vspace{-5pt}
\end{wrapfigure}
We implement a gradient accumulation strategy that updates the gradients every 10 steps. The batch of images is drawn from the auxiliary dataset $\mathcal{D}_{aux}$ and is used to assess the performance of the target network constructed based on the selected architecture.
HyperNAS undergoes training for 200 epochs, employing the ADAM optimizer with an initial learning rate of 1e-3, which is halved at epochs 100 and 150. It's important to note that batch statistics in Batch Normalization (BN) operations are computed directly by the PyTorch framework, rather than being predicted by the hypernetwork. 
Given the fixed output dimension of the hypernetwork, the output of the hypernetwork is reshaped and sliced based on the parameter shapes in each node.
Additionally, to maintain a balance between the dual tasks within HyperNAS, we set the preference factor q to 1.5.\\
\noindent\textbf{GCN encoder settings.} The shared GCN encoders in HyperNAS are configured differently depending on the search space. For the DARTS~\cite{liu2018darts} search space, the encoder has 3 layers with a hidden dimension of 144. For the ViT search space, it also has 3 layers but with a hidden dimension of 256. For other search spaces~\cite{ying2019bench,dong2020bench,howard2019searching}, the encoder consists of 6 layers with a hidden dimension of 72.\\
\noindent\textbf{Hypernetwork settings.} In HyperNAS, the architecture of the shared hypernetworks is customized according to the specific search space. For the DARTS~\cite{liu2018darts} search space, the hypernetwork is designed with 2 layers, each having a hidden dimension of 128. For ViT, the hypernetwork comprises 4 layers, each also featuring a hidden dimension of 128. Conversely, for other search spaces~\cite{ying2019bench,dong2020bench,howard2019searching}, the hypernetwork is configured with 3 layers, each with a hidden dimension of 96.

\section{Search Spaces}
\noindent\textbf{NAS benchmarks.} NAS-Bench-101~\cite{ying2019bench} and NAS-Bench-201~\cite{dong2020bench} are widely used benchmarks in the NAS domain. NAS-Bench-101 contains 423,624 unique architectures, each constructed with 9 repeated cells. The cell within these architectures can contain up to 7 nodes and 9 edges. NAS-Bench-201 contains 15,625 unique architectures, all featuring a consistent macro structure of 15 repeated cells. Each cell is comprised of 4 nodes connected by up to 6 edges. \vspace{5pt}\\
\noindent\textbf{DARTS.} DARTS~\cite{liu2018darts} is an open-domain search space designed for NAS. It constructs architectures using both normal and reduction cells. Each cell consists of 7 nodes and 14 edges, where each edge connects to one of 7 candidate operations. \vspace{5pt}\\
\noindent\textbf{MobileNetV3.} MobileNetV3 consists of 5 stages, with each stage containing 2 to 4 building blocks. This results in a maximum of 20 layers (5 stages × 4 blocks per stage). Each building block offers flexibility in its configuration: the kernel size is selected from \{3, 5, 7\}, while the expansion ratio is chosen from \{3, 4, 6\}. \vspace{5pt}\\
\noindent\textbf{ViT.} Following Autoformer, we use a search space that encompasses five key architectural factors in transformer blocks: embedding dimension, Q-K-V dimension, number of heads, MLP ratio, and network depth. The detailed configuration is summarized in Table \ref{table:vit-sp}.

\begin{table}[b]
\tablestyle{12pt}{1.3}
\caption{Search results on NAS-Bench-201. Three runs are conducted with different random seeds. }
\label{table:nas201}
\begin{tabular}{ccccccc}
\hline
 & \multicolumn{2}{c}{CIFAR-10} & \multicolumn{2}{c}{CIFAR-100} & \multicolumn{2}{c}{ImageNet-16-120} \\ \cline{2-7} 
\multirow{-2}{*}{Method} & validation & test & validation & test & validation & test \\ \hline
PINAT & 91.03±0.54 & 94.09±0.14 & 73.17±0.31 & 73.26±0.19 & 45.95±0.21 & \textbf{46.55±0.72} \\ 
\rowcolor[HTML]{EFF6FB} 
\textbf{HyperNAS} & \textbf{91.57±0.00} & \textbf{94.37±0.00} & \textbf{73.43±0.01} & \textbf{73.40±0.03} & \textbf{46.65±0.02} & 46.50±0.00 \\ \hline
\end{tabular}
\end{table}

\section{Further Experiments}
\subsection{Search Results on NAS-Bench-201}
To further evaluate the effectiveness of HyperNAS, we conducted experiments to search for optimal architectures on NAS-Bench-201. Experiments are performed with three independent runs, with the predictor trained on 156 training samples. We use the current state-of-the-art predictor, PINAT, as our baseline for comparison. The search results, shown in Table \ref{table:nas201}, demonstrate that HyperNAS identifies superior architectures and exhibits strong robustness across all datasets.

\begin{table}[t!]
  \begin{minipage}{0.48\linewidth}
 \vspace{-12pt}
    \centering
    \captionsetup{type=table}
    \caption{Performance comparison of architectures searched by SOTA methods on ImageNet in the MobileNetV3 search space.}
    \label{table:mobileNetV3}
    \tablestyle{5pt}{1.2}
    \begin{tabular}{cccccc}
        \hline
        Method & 
        \begin{tabular}{@{}c@{}}Param\\(M)\end{tabular} & 
        \begin{tabular}{@{}c@{}}FLOPs\\(M)\end{tabular} & 
        \begin{tabular}{@{}c@{}}Top-1\\(\%)\end{tabular} & 
        \begin{tabular}{@{}c@{}}Top-5\\(\%)\end{tabular} & 
        \begin{tabular}{@{}c@{}}Queries\\(K)\end{tabular} \\ \hline
        \multicolumn{6}{c}{MobileNetV3-like} \\ \hline
        MobileNetV3 & 5.4 & 219 & 75.2 & - & - \\
        FBNet-C & 5.5 & 375 & 74.9 & - & 11.5 \\
        MnasNet-A3 & 5.2 & 403 & 76.7 & 93.3 & 8 \\
        ProxylessNAS & 7.1 & 465 & 75.1 & 92.3 & - \\
        OFA & - & 230 & 76.0 & - & 16 \\
        SPOS & 5.4 & 472 & 74.8 & - & - \\
        RLNAS & 5.3 & 473 & 75.6 & 92.6 & - \\
        NP & 6.4 & 536 & 74.8 & - & - \\
        FBNetV2 & - & 321 & 76.3 & 92.9 & 11.5 \\
        AtomNAS & - & 367 & 75.9 & 92.0 & 78 \\
        CTNAS & - & 482 & 77.3 & 93.4 & 1 \\
        PINAT & 5.1 & 452 & 77.8 & 93.5 & - \\ 
        \rowcolor[HTML]{EFF6FB} 
        \textbf{HyperNAS} & 5.7 & 428 & \textbf{77.9} & \textbf{93.6} & \textbf{0.2} \\ \hline
    \end{tabular}
    \end{minipage}
    \hspace{3mm}
    \begin{minipage}{0.48\linewidth}
    \caption{Comparison results for cell connectivity strategy. }
    \label{table:cell_connect}
    \tablestyle{2pt}{1.4}
    \begin{tabular}{ccccc}
    \hline
    Connectivity& 50 (0.01\%)& 100 (0.02\%) & 172 (0.04)\%) & 424 (0.1)\%\\\hline
    Concatenate  & 0.675 & 0.684 & 0.714 & 0.778\\
    \rowcolor[HTML]{EFF6FB} 
    Sum & \textbf{0.678} & \textbf{0.688} & \textbf{0.736} & \textbf{0.779} \\ \hline
    \end{tabular}

    \caption{
    Training cost across different search spaces, measured in GPU days. ViT is trained on \textbf{ImageNet}, while the others are trained on CIFAR-10.}
    \label{table:nas101}
    \tablestyle{8pt}{1.4}
    \begin{tabular}{c|cccc}
    \hline
    Search space& NB101& NB201 & DARTS & ViT\\\hline
    Training time  & 0.3 & 0.2 & 0.3 & 2\\
    \hline
    \end{tabular}

    \tablestyle{3.pt}{1.4}
    \caption{Time cost of evaluating a single architecture in NAS-Bench-101.}
    \label{tab:time_cost}
    \begin{tabular}{cccccc}
    \hline
    \multicolumn{1}{c|}{Method} & NP & GHN & PINAT & HyperNAS & Full training \\ \hline
    \multicolumn{1}{c|}{Time cost} & 0.002s & 14s & 0.012s & 0.014s & 8h \\ \hline
    \multicolumn{1}{l}{} & \multicolumn{1}{l}{} & \multicolumn{1}{l}{} & \multicolumn{1}{l}{} & \multicolumn{1}{l}{} & \multicolumn{1}{l}{}
    \end{tabular}

\end{minipage}
\vspace{-10pt}
\end{table}

\subsection{Search Results on MobileNetV3}
\label{sec:mobilenet}
For MobileNetV3, we represent each architecture with five normal cells, considering each stage as a cell. A shared GCN encoder is employed to encode the various cells. Moreover, we evaluate the searched architecture using the strategy from CTNAS~\cite{chen2021contrastive} and restrict the number of FLOPs to less than 500M. As shown in Table \ref{table:mobileNetV3}, HyperNAS achieves the highest accuracy with just 200 queries, a query count that is orders of magnitude fewer than other methods. These outcomes further emphasize the robustness of HyperNAS in few-shot scenarios.

\begin{figure}[b]
    \centering
    \captionsetup[subfigure]{margin=2pt}
    \begin{minipage}[b]{0.49\textwidth}
        \centering
        \subfloat[Ablation studies on ranking abilities of the neural predictor.]{
            \label{subfig:pred_201}
            \includegraphics[width=0.47\linewidth]{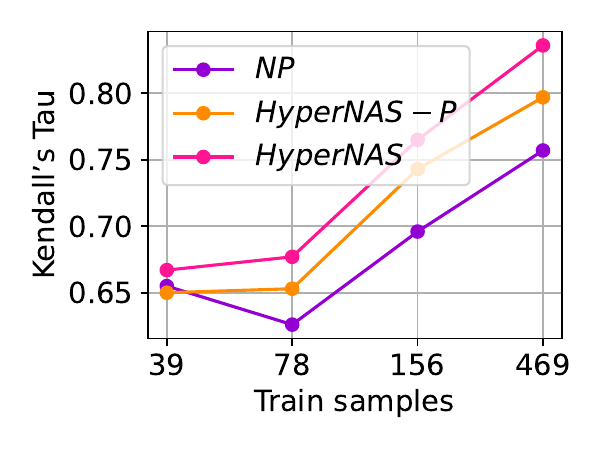}
        }
        \subfloat[Ablation studies on parameter prediction of the hypernetwork.]{
            \label{subfig:hyper_201}
            \includegraphics[width=0.49\linewidth]{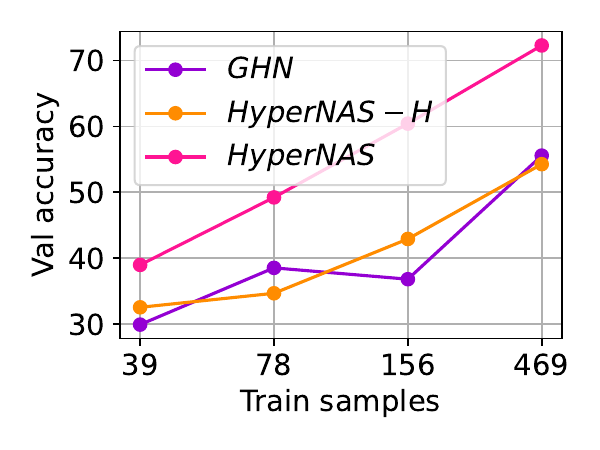}
        }
        \caption{Ablation studies on NAS-Bench-201 search space.}
        \label{fig:ablate_predictor}
    \end{minipage}
    \hspace{2pt}
    \begin{minipage}[b]{0.47\textwidth}
        \centering
        \subfloat[Ranking abilities of the neural predictor.]{
            \label{subfig:q_predictor_101}
            \includegraphics[width=0.5\linewidth]{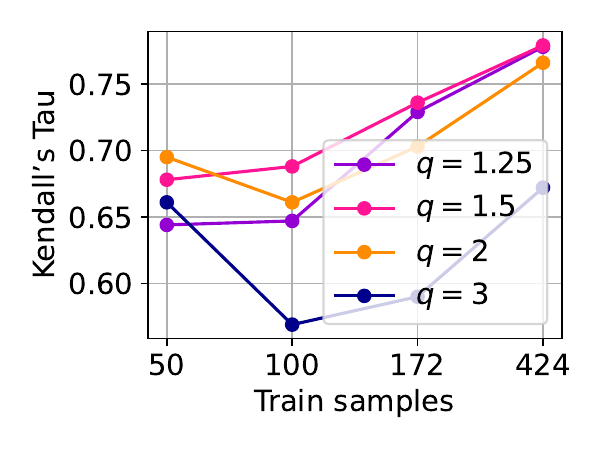}
        }
        \subfloat[Parameter prediction capabilities of the hypernetwork.]{
            \label{subfig:q_hypernet_101}
            \includegraphics[width=0.51\linewidth]{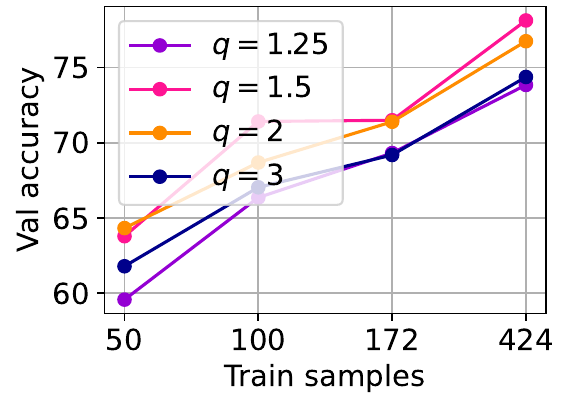}
        }
        \caption{Comparison results of adaptive multi-task loss with varying $q$ values.}
        \label{fig:ablate_q}
    \end{minipage}
\end{figure}

\subsection{Ablation Studies}
\noindent\textbf{Ablation studies on NAS-Bench-201.} Figure \ref{fig:ablate_predictor} presents the ablation results on NAS-Bench-201~\cite{dong2020bench}, showing similar trends as those on NAS-Bench-101~\cite{ying2019bench}. As depicted in Figure \ref{subfig:pred_201}, HyperNAS-P significantly improves ranking abilities compared to NP, while HyperNAS explicitly outperforms HyperNAS-P. These improvements demonstrate the effectiveness of the proposed global encoding scheme and the shared hypernetwork. When compared to single-task HyperNAS-P and HyperNAS-H, HyperNAS exhibits superior performance in both tasks, as illustrated in Figures \ref{subfig:pred_201} and \ref{subfig:hyper_201}. These findings underscore the compatibility between neural predictors and the shared hypernetwork, as well as the effectiveness of our multi-task loss function. \vspace{5pt}\\
\noindent\textbf{Effect of GCN encoder on the hypernetwork.} 
To assess the impact of the GCN encoder on the hypernetwork, we conducted experiments on GHN~\cite{zhang2018graph} and HyperNAS-H. GHN uses a GNN encoder with forward-backward propagation, whereas HyperNAS-H employs a specially designed GCN encoder. Figure \ref{subfig:hyper_101} shows that HyperNAS-H consistently achieves higher validation accuracy than GHN, demonstrating the superior capability of our encoder in capturing architectural information. \vspace{5pt}\\
\noindent\textbf{Effect of cell connectivity.} Table \ref{table:cell_connect} presents the ablation studies of cell connectivity. It can be seen the performance of the concatenate strategy is slightly weaker than the sum strategy, showcasing the strength of the sum strategy in capturing the macro-structure of architectures. Additionally, the sum strategy is more parameter-efficient, making it a preferable choice. \vspace{5pt}\\
\noindent\textbf{Impact of $q$ value on multi-task balance.} 
The $q$ value plays a crucial role in our proposed dynamic adaptive multi-task training objectives, facilitating personalized exploration of solutions on the Pareto front. To intuitively investigate its effectiveness, we conducted experiments with various q values $\{1.25, 1.5, 2, 3\}$. Notably, when $q=2$, the loss function aligns with an adaptive multi-task loss from previous work~\cite{liebel2018auxiliary}. As depicted in Figure \ref{fig:ablate_q}, both tasks achieve optimal performance at $q=1.5$ across most data splits, surpassing the conventional adaptive loss function with $q=2$. This highlights the advancement and flexibility of our adaptive multi-task loss function.


\subsection{Efficiency Evaluation}
\noindent\textbf{Training cost.} Table \ref{table:nas101} reports training costs across different search spaces. MobileNetV3 is trained on ImageNet-1K, while other search spaces are tested on the CIFAR10. The results show that resource consumption remains within an affordable range, including for ViT on ImageNet-1K. \\
\noindent\textbf{Evaluation cost.} Table \ref{tab:time_cost} presents the time cost comparison for evaluating a single neural architecture in NAS-Bench-101. Notably, HyperNAS achieves a time cost 1000 times lower than GHN which directly employs hypernetworks for architecture evaluation, highlighting the high efficiency of our proposed paradigm.  Moreover, HyperNAS achieves comparable efficiency to the current state-of-the-art predictor PINAT, indicating its practical feasibility. 


\begin{figure}[b!]
    \centering
    \captionsetup[subfigure]{margin=0pt}
    \subfloat[Normal cell.]{
        \label{subfig:q_predictor_101}
        \centering
    \includegraphics[width=0.5\linewidth]{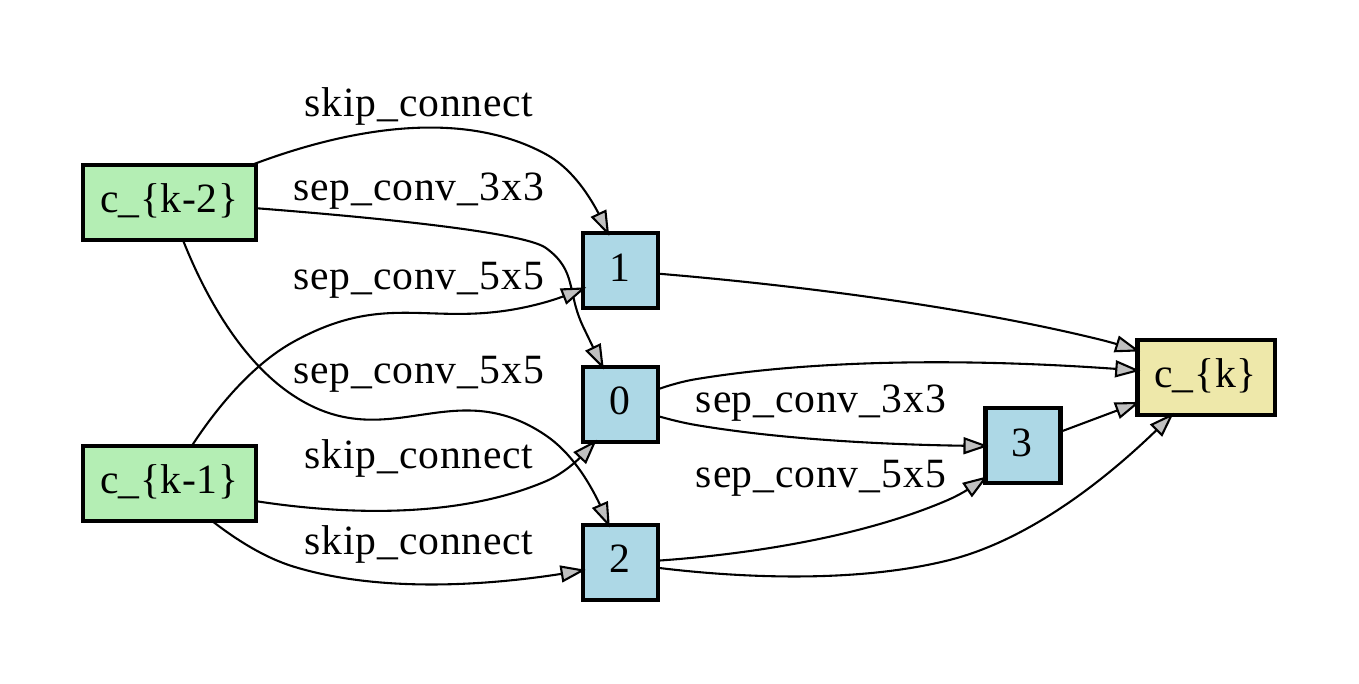}}
    \subfloat[Reduction cell.]{
        \label{subfig:q_hypernet_101}
        \centering     
        \includegraphics[width=0.5\linewidth]{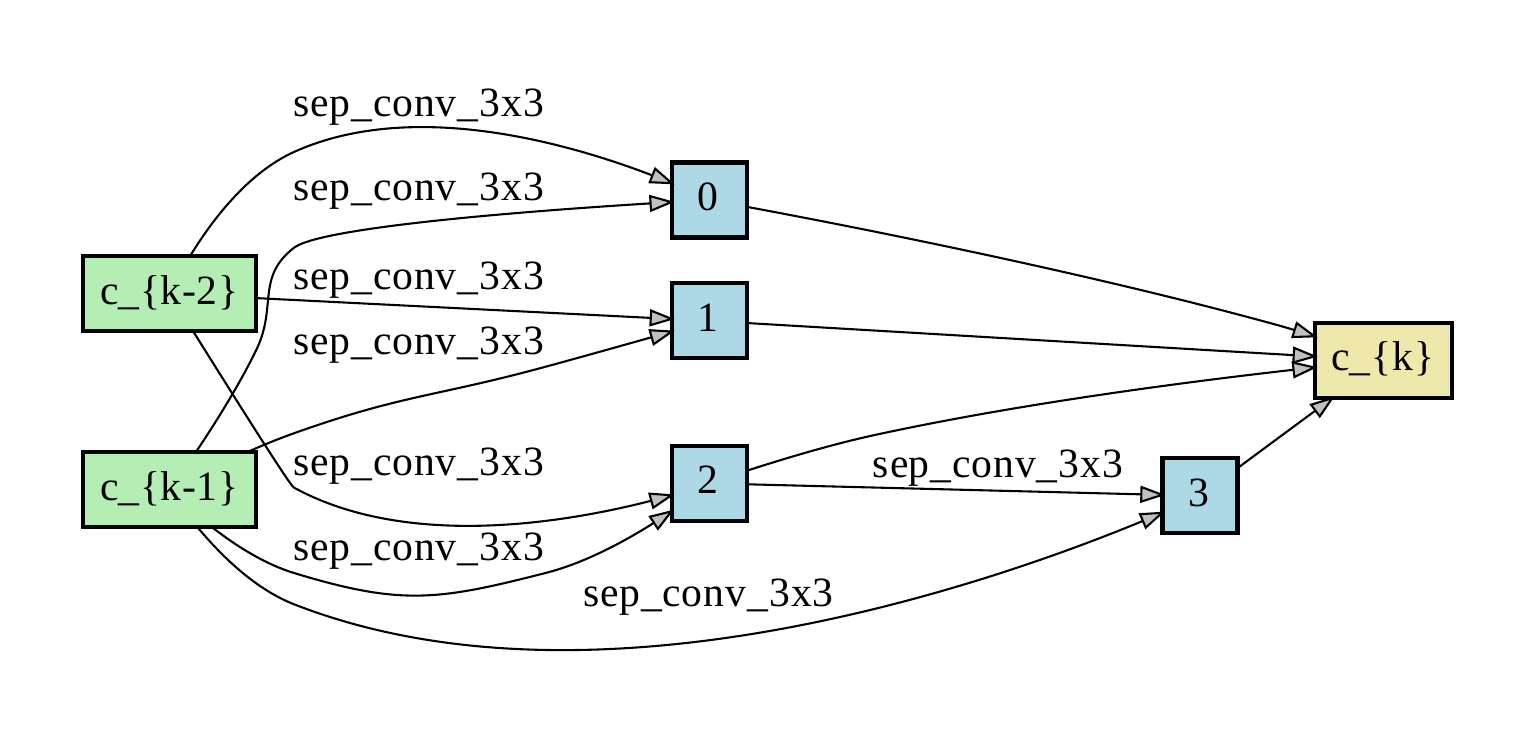}
        }
    \caption{
The architecture of the convolutional cells found by HyperNAS-P in the DARTS search space.}
    \label{fig:arch-HyperNAS-P}
\end{figure}

\begin{figure}[t!]
    \centering
    \captionsetup[subfigure]{margin=0pt}
    \subfloat[Normal cell.]{
        \label{subfig:q_predictor_101}
        \centering
    \includegraphics[width=0.5\linewidth]{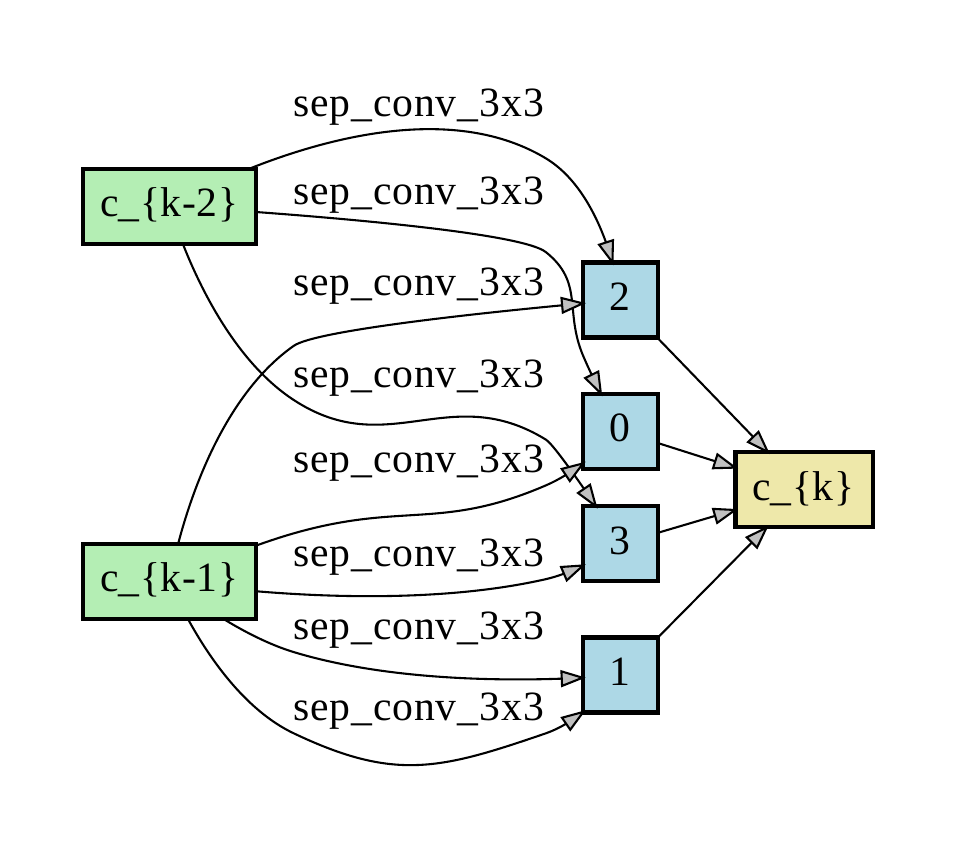}}
    \subfloat[Reduction cell.]{
        \label{subfig:q_hypernet_101}
        \centering     
        \includegraphics[width=0.5\linewidth]{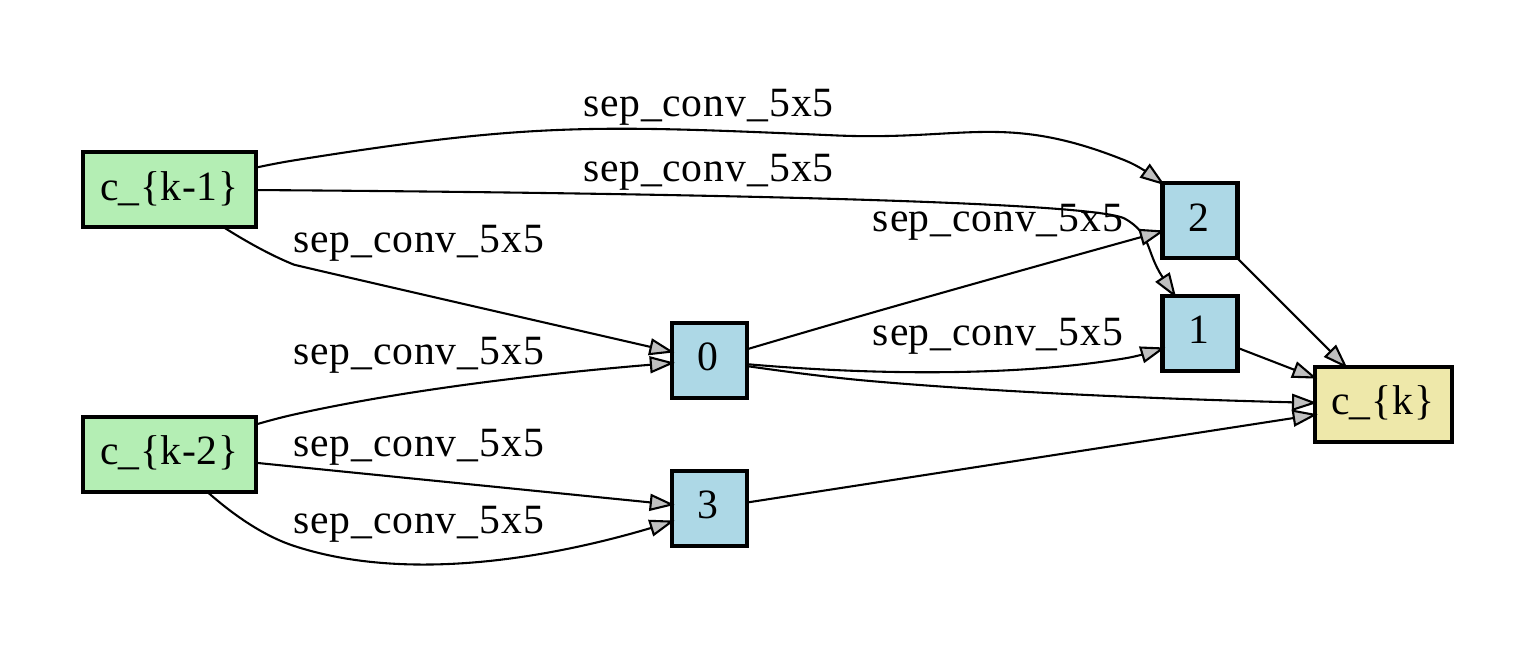}
        }
    \caption{The architecture of the convolutional cells found by HyperNAS in the DARTS search space.}
    \label{fig:arch-HyperNAS}
\end{figure}

\section{Visualization of Searched Architectures}
\begin{figure}[h!]
  \centering
  \includegraphics[width=0.85\textwidth]{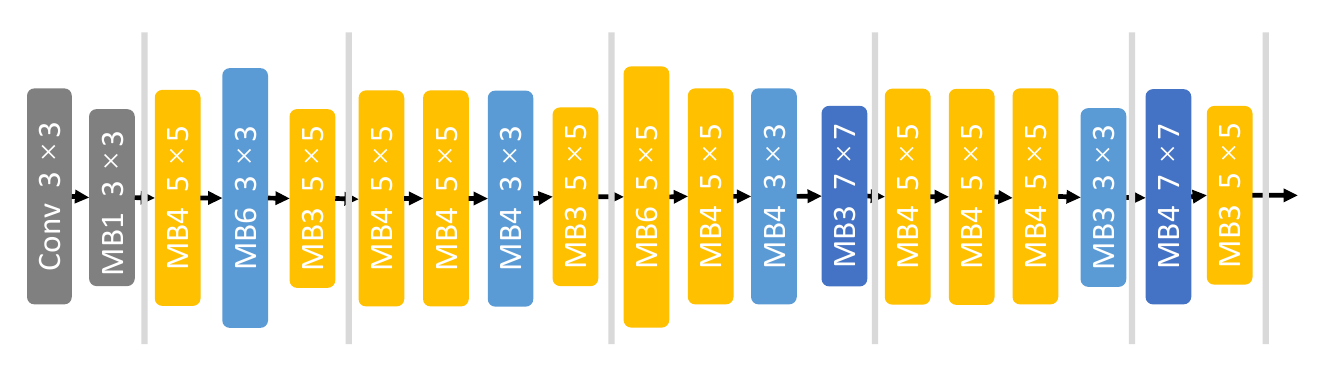}
  \caption{The architecture found by HyperNAS in the MobileNetV3 search space on ImageNet.}
   \label{fig:arch-MBV3}
\end{figure}

\noindent\textbf{DARTS.} Figures \ref{fig:arch-HyperNAS-P} and \ref{fig:arch-HyperNAS} display architectures found by HyperNAS-P and HyperNAS on CIFAR-10, which achieve top-1 accuracy of 97.61\% and 97.60\% on CIFAR-10, respectively. Interestingly, in the architecture discovered by HyperNAS (Figure \ref{fig:arch-HyperNAS}), all operations within the normal cell are ``sep\_conv\_3$\times$3'', and all within the reduction cell are ``sep\_conv\_5$\times$5'', with no parameter-free operations (e.g., skip connection or max pooling). 
By comparing with the architecture found by HyperNAS-P (Figure \ref{fig:arch-HyperNAS-P}), this phenomenon can be attributed to the shared hypernetwork in HyperNAS, which is employed to generate weights for various architectures.\\
\noindent\textbf{MobileNetV3.} Figure \ref{fig:arch-MBV3} displays the architecture found by HyperNAS in the MobileNet-v3 search space on ImageNet~\cite{krizhevsky2017imagenet}. \\

\noindent\textbf{ViT.} Table \ref{table:arch-vit} displays the architecture found by HyperNAS in the ViT search space on ImageNet~\cite{krizhevsky2017imagenet}.
\begin{table}[h]
\caption{The architectures found by HyperNAS in the ViT search space on ImageNet.}
\label{table:arch-vit}
\tablestyle{2pt}{1.2}
\begin{tabular}{ccccc}
\hline
method & depth & mlp ratio & head nums & embead dim \\ \hline
HyperNAS-s & 13 & {[}3.5, 4.0, 3.5, 4.0, 4.0, 3.5, 4.0, 4.0, 4.0, 4.0, 4.0, 3.5, 3.5{]} & {[}5, 5, 7, 6, 6, 5, 7, 6, 5, 7, 6, 5, 6{]} & 384 \\
HyperNAS-b & 14 & {[}3.0, 3.5, 3.5, 3.5, 4.0, 3.5, 4.0, 3.0, 3.5, 4.0, 4.0, 3.5, 4.0, 3.5{]} & {[}9, 9, 9, 10, 9, 9, 10, 10, 10, 10, 9, 9, 10, 10{]} & 576 \\ \hline
\end{tabular}
\end{table}

\end{document}